%% file: main_arxiv.tex
\documentclass[pdflatex,sn-mathphys-num]{sn-jnl}

\usepackage{graphicx}%
\usepackage{multirow}%
\usepackage{amsmath,amssymb,amsfonts}%
\usepackage{amsthm}%
\usepackage{mathrsfs}%
\usepackage[title]{appendix}%
\usepackage{xcolor}%
\usepackage{textcomp}%
\usepackage{manyfoot}%
\usepackage{booktabs}%
\usepackage{algorithm}%
\usepackage{algorithmicx}%
\usepackage{algpseudocode}%
\usepackage{listings}%

\theoremstyle{thmstyleone}%
\theoremstyle{thmstyletwo}%
\newtheorem{remark}{Remark}%

\theoremstyle{thmstylethree}%
\newtheorem{definition}{Definition}%

\usepackage{lineno}
\input{prefix}

\begin{document}

\def\entitle{Algorithmic Fairness: Not a Purely Technical but Socio-Technical Property}
\title[]{\entitle}

\author*[1]{\fnm{Yijun} \sur{Bian}}\email{yibi@di.ku.dk}
\author[2]{\fnm{Lei} \sur{You}}\email{leiyo@dtu.dk}
\author[3]{\fnm{Yuya} \sur{Sasaki}\email{sasaki@ist.osaka-u.ac.jp}}
\author[4]{\fnm{Haruka} \sur{Maeda}\email{maeda.haruka.5j@kyoto-u.ac.jp}}
\author[5]{\fnm{Akira} \sur{Igarashi}\email{igarashi.a.hus@osaka-u.ac.jp}}

\affil[1]{\orgdiv{Department of Computer Science}, \orgname{University of Copenhagen}, \orgaddress{\postcode{2100} \city{København Ø}, \country{Denmark}}}
\affil[2]{\orgdiv{Department of Engineering Technology}, \orgname{Technical University of Denmark}, \orgaddress{\postcode{2750} \city{Ballerup}, \country{Denmark}}}
\affil[3]{\orgdiv{Graduate School of Information Science and Technology}, \orgname{The University of Osaka}, \orgaddress{\postcode{565-0871}, \state{Osaka}, \country{Japan}}}
\affil[4]{\orgdiv{Graduate School of Law}, \orgname{Kyoto University}, \orgaddress{\postcode{606-8501}, \state{Kyoto}, \country{Japan}}}
\affil[5]{\orgdiv{Department of Human Sciences}, \orgname{The
University of Osaka}, \orgaddress{\postcode{565-0871}, \state{Osaka}, \country{Japan}}}




\abstract{

The rapid trend of deploying artificial intelligence (AI) and machine learning (ML) systems in socially consequential domains has raised growing concerns about their trustworthiness, including potential discriminatory behaviours. Research in algorithmic fairness has generated a proliferation of mathematical definitions and metrics, yet persistent misconceptions and limitations---both within and beyond the fairness community---limit their effectiveness, such as an unreached consensus on its understanding, prevailing measures primarily tailored to binary group settings, and superficial handling for intersectional contexts. 
Here we critically remark on these misconceptions and argue that fairness cannot be reduced to purely technical constraints on models; we also examine the limitations of existing fairness measures through conceptual analysis and empirical illustrations, showing their limited applicability in the face of complex real-world scenarios, challenging prevailing views on the incompatibility between accuracy and fairness as well as that among fairness measures themselves, and outlining three worth-considering principles in the design of fairness measures. 
We believe these findings will help bridge the gap between technical formalisation and social realities and meet the challenges of real-world AI/ML deployment.

}


\maketitle

\section{Introduction}
\label{sec1}

Due to an increasing trend of deploying artificial intelligence (AI) systems or machine learning (ML) models in high-stakes domains such as healthcare, recruitment, and jurisdiction, there have been raised concerns regarding their trustworthiness, including whether underlying discrimination is hidden there \citep{obermeyer2019dissecting,vlasceanu2022propagation,chen2023algorithmic,hu2024generative,glickman2024human,jones2024causal}. Research in algorithmic fairness (or fairness in machine learning, FairML) seeks to prevent ML algorithms or models from exacerbating improper human prejudices, with a further aim to contribute to building a more equitable society. Despite the substantial progress that has been made in this field, misconceptions about the scope and meaning of fairness persist, both within and outside the FairML community. 
For instance, fairness is too often realised as a purely mathematical constraint on learning algorithms or ML models, detached from the social realities from which the issue originates. This framing not only obscures the role of structural inequalities, but also encourages the prejudice solidified in technical fixes, against our premier will to address discrimination in practice. At the same time, many widely used fairness measures remain grounded in one protected (or sensitive) attribute that is usually binary, systematically underestimating disparities when sensitive attributes are multi-valued or intersectional. Incremental extensions---through binarisation or traversal-based generalisations---either oversimplify or introduce prohibitive computational costs, leaving fairness assessments poorly aligned with real-world needs in mitigating discrimination.

In this perspective, we first clarify and discuss several confusing and recurring misconceptions about fairness research, especially those that exist outside this field. 
We then summarise existing fairness measures in the literature and present some empirical results to show why they fall short in cases of non-binary sensitive attributes. 
In what follows, we rethink and challenge a few prevailing views in this field (such as the conflict between accuracy and fairness, and the incompatibility between group fairness and individual fairness) by empirical illustrations. 
A detailed survey of existing fairness definitions and measures, as well as our experimental setups and additional empirical results, is provided in the appendices. 
Building on these observations, we propose guiding principles for the design of reasonable fairness measures that are both conceptually rigorous and practically viable, in order to advance relevant research. In doing so, we aim to reframe how fairness itself is understood in ML and bridge the gap between technical definitions and societal realities, and to chart a path toward fairness measures that scale with the real-world complexity and societal impact of contemporary AI systems.

\section{Confusing concepts: Distinguishing discrimination from unfairness}

Misunderstandings often emerge in discussions between FairML researchers and scholars in AI/ML who do not specialise in fairness. Even within fairness-relevant studies, the term \emph{fairness} may denote distinct concepts. For instance, the fairness in \citet{rampisela2025joint} refers to how evenly the resources are distributed to items in recommendation systems, while much of the literature focuses on the differential treatment of individuals by people or machines \citep{binns2018fairness,caton2020fairness, mitchell2021algorithmic, mehrabi2021survey, pessach2022review, ferrara2023fairness}. 
Other scholars question whether fairness is even a resolvable issue, citing examples such as: ``Is it unfair that a banker’s salary differs from that of a security guard? Is it unfair that incomes vary across different countries? What if people are discriminated against by their food preferences?''

These difficulties in communication across research areas partly derive from the interchangeable use of terms such as \emph{fairness, unfairness, discrimination}, and \emph{bias} in the FairML literature; however, they may actually mean something different, especially in the context of social sciences. 
In social science, unfairness denotes uneven resource allocation that may not necessarily be caused by discrimination---such as the discrepancy in incomes between high- and low-income countries, or salary differences between job levels---where differences arise from context or merit instead of immutable characteristics. Bias, meanwhile, may refer to deviations arising from sampling or equipment errors, and therefore does not always equate to discrimination. 
An analogous example is the subtle difference between equality and equity: everyone gets exactly the same in social justice (equality), while in distributive justice (equity), those with more need would get the amount of resources they need, with common instantiations of some resources used to encourage the groups that are marginalised or underrepresented in history.

\emph{The FairML research originated from discrimination issues in reality tied to certain protected attributes (such as race, gender, age, disability, pregnancy, and sexual orientation)} in contexts like university admissions in education, recidivism risk assessment in judicature, and medical provision in healthcare. 
For such scenarios, one of the obstacles is the lack of clean data because the observed data from reality are often already biassed. It means that: fully relying on data or solely mathematical formulations of fairness can result in a misalignment with societal needs: they risk privileging correlation over causation, enabling ``fair in form but biassed in effect'' outcomes (\aka{} differential effect in the context of discrimination in law), and ignoring the systemic inequities that generate disparities in the first place.

\section{Misconceptions outside the field: Why a purely mathematical framing of fairness is inadequate}

Discussions with scholars outside the FairML community reveal several recurring viewpoints that differ from the prevailing consensus within the field. 
\begin{itemize}
\item \textbf{Fairness through unawareness or separation}:  
A common question often raised immediately is that, by omitting sensitive attributes from the data and allowing a model to learn without them, we can achieve fairness and avoid the need for explicit interventions or more sophisticated mechanisms. Another related claim suggests that, since data imbalance may disadvantage marginalised groups, while improving fairness could reduce accuracy for the privileged group, it would be ``fairer'' to train separate models for each subgroup.

\item \textbf{Fairness beyond protected attributes}: 
Some contend that fairness should apply to any attribute, not only those legally or socially recognised as protected. They point to possible discrimination based on attributes like eye or hair colour, or even dietary preference, on the basis that such distinctions ``simply can be made''.

\item \textbf{Ambiguity in identifying privileged groups}: 
Others note that given a protected attribute, the privileged group may not be uniquely defined, and more than one group can be relatively advantaged depending on the context. 
For example, median income statistics in the United States indicate that Asian individuals earn more on average than white individuals, suggesting context-dependent privilege: Asians should be considered privileged in this specific setting.

\item \textbf{Mathematical vs. social remit}: 
A further view holds that determining who is privileged or disadvantaged lies outside the domain of computer scientists. Under this position, fairness should be defined in purely mathematical terms, applied universally, and leave social and ethical questions to other disciplines.
\end{itemize}

We suggest that these perspectives, while often well-intentioned, overlook several crucial conceptual and practical considerations. 

First, simply removing protected attributes from the data will not ensure fairness or eliminate discrimination, because information correlated with protected characteristics can be encoded in some non-protected features that serve as proxies. 
Likewise, it risks reinforcing entrenched disparities rather than mitigating them to stratify individuals by group membership and develop separate models for each subgroup, particularly when subgroup definitions themselves reflect underlying structural inequalities, such as stereotypical women or certain ethnic groups, reflected by ``justifiable'' forms that are easily morphed into from discrimination starting as taste-based or inefficient \citep{bertrand2017field}.

Second, extending fairness to all attributes requires careful differentiation. Outcome differences tied to voluntary choices or role-specific factors---such as dietary preferences or salary differences between academic ranks---are not equivalent to disparities rooted in structural bias. Furthermore, attributes appearing unrelated to protected characteristics (\eg{} hair colour, educational attainment) often function as proxies, carrying implicit links to existing inequities. Identifying and addressing such proxy attributes is also essential to avoid misdiagnosing the source of disparity.

Third, statistical disparities between groups alone do not establish causal determination associated with protected attributes. In Denmark, for example, higher salaries among foreign workers may be the result of minimum salary thresholds imposed by immigration policy, rather than labour market advantage per se. 
Recognising such confounding factors is critical to avoid erroneous conclusions, including incorrect identification of privileged groups. While privilege can indeed be context-dependent, beginning with a simplified assumption---such as one single privileged group---can still offer tractable insights before extending to more complex, multi-privilege settings.

Finally, invoking a lack of social science expertise as a justified reason to disengage is unsound and problematic. In research more broadly, unfamiliarity with a relevant method is not an acceptable basis for exclusion. 
By analogy, if FairML intersects directly with racial or gender discrimination, understanding these relevant social contexts is part of the responsibility of those addressing it. Without this grounding, a ``solution'' may address an artefact of the data rather than the real-world problem.

In summary, FairML is inherently socio-technical, and both rigorous mathematical formulation and informed engagement with the social realities in which these systems operate are required to address it effectively. 
Treating ethical considerations as external to the technical process risks producing ML systems that are formally ``fair'' but substantively misaligned with societal needs.

\section{Underestimation within the field: Why existing fairness measures remain insufficient}

A rich set of formal definitions and statistical criteria has been developed in existing fairness research, summarised in Table~\ref{tab:fair,summary} (see Appendix~\ref{sec:related} for details). 
Given that, some scholars value more research on providing normative ML pipelines to assist researchers in making decisions rather than proposing new computational methods for fairness.
Despite the important starting points they offer, these criteria remain limited in scope. 
Below, we highlight several conceptual and technical issues that arise when such measures are applied to contemporary ML practice, with experimental setups and additional results detailed in Appendix~\ref{sec:expert} to \ref{sec:appx}.

\subsection{Statistical criteria and their limitations}
Three canonical criteria---independence, separation, and sufficiency---are widely recognised as cornerstones of formal definitions of statistical non-discrimination \citep{barocas2023fairness}. Respectively, they correspond to requiring predictions to be independent of protected attributes ($\mathsf{A \indep R}$), conditionally independent given outcomes ($\mathsf{R \indep A \mid Y}$), or that true labels be independent of protected attributes given predictions ($\mathsf{Y \indep A \mid R}$). 
These conditions are rarely achievable simultaneously except in trivial settings, and even when satisfied, may yield unintended consequences. For instance, strict independence can generate perverse outcomes, such as the so-called glass cliff: equal hiring rates across groups may still disadvantage one group if selection quality differs due to structural imbalances in training data. 
Furthermore, it is difficult for discrimination in statistical models to effectively account for the multitude of factors relevant to unequal outcomes, such as a same level of assertiveness being deemed as good for productivity when coming from men but bad when coming from women \citep{bertrand2017field}, and vicious circles due to unconscious bias and self-fulfilling prophecies. 
This becomes a main liability of this approach and leaves open the possibility that the disparities attributed to discrimination may in fact be explained by some other unmeasured cause(s), and therefore should be used cautiously in making causal interpretations \citep{pager2008sociology}.
Such issues illustrate that meeting a statistical condition is not equivalent to ensuring substantive fairness.

In addition, existing fairness measures following such criteria, known as group fairness, are often defined for one (typically binary) protected attribute only, which limits their applicability in reality where multi-valued and intersectional attributes exist. Some scholars view it as straightforward to do traversal-based generalisation for one multi-valued \senatt{} and even intersectional attributes, unaware of how much extra time it would cost compared to their original definitions. 
In the following, we list several commonly used group fairness measures, originally defined for one bi-valued protected attribute (\aka{} sensitive attribute), and extend them to apply to non-binary cases.
Note that a task relevant to discrimination or bias mitigation is usually binary classification or prediction using a score function in some cases, and for instance $(\xneg,a_1)$ where $a_1$ denotes one bi-valued \senatt{} (\ie{} $a_1\in \mathcal{A}_1=\{0,1\}$), $a_1=1$ represents the privileged group, while $a_1=0$ represents the marginalised group. 
Notation $\probP$ is the probability measure.

\paragraph{Independence-based measures of fairness}

Demographic parity \citep{gajane2017formalizing,jiang2020wasserstein} (DP, \aka{} statistical parity \citep{dwork2012fairness,chouldechova2017fair}), that is,
\begin{equation}
\topequation
\mathrm{DP}(f)= |
    \probP( f(\xneg,a_1)=1 \mid a_1=0 )-
    \probP( f(\xneg,a_1)=1 \mid a_1=1 )
| \,,\label{eqn,grp1}
\end{equation}
along with related notions (such as disparate impact, disparate treatment, conditional statistical parity, and bounded group loss), instantiates the independence criterion. 
Note that disparate treatment \citep{zafar2017fairness2} is also indicated as ``statistical parity'' \citep{corbett2017algorithmic,haas2019price} and defined as 
$\probP( f(\xneg,a_1)=1 \mid a_1=j )
    = \probP( f(\xneg,a_1)=1 )
    ,\, \forall\, j\in\{0,1\}
    .\,$ 
While mathematically straightforward, DP is limited in scope because its canonical form applies only to one single binary \senatt{}. 
Extensions to multi-valued attributes are often achieved through binarisation, grouping all non-privileged categories together. 
That is to say, it can be evaluated by
\begin{equation}
\topequation
\mathrm{DP}'(f) =|
    \probP( f(\xneg,a_1)=1 \mid a_1\neq 1 )-
    \probP( f(\xneg,a_1)=1 \mid a_1=1 )
| \,,\label{eqn,grp1,bin}
\end{equation}
where $a_1\in\mathcal{A}_1=\{1,2,...,n_{a_1}\}$, $a_1=1$ still denotes the privileged group, and $a_1\neq 1$ denotes the marginalised groups. 
Note that \eqref{eqn,grp1,bin} is equivalent to \eqref{eqn,grp1} when $\mathcal{A}_1=\{0,1\}$.

However, binarisation oversimplifies the discrimination complexity in reality and risks masking important disparities: for example, treating several marginalised groups as homogeneous may obscure intersectional harms. 
One refined extension, inspired by statistical parity \citep{corbett2017algorithmic,agarwal2019fair} (or statistical parity difference \citep{chen2024fairness}), can be defined as
\begin{equation}
\topequation
\mathrm{DP}^\text{ext}(f) =
\textstyle \max_{ j\in\mathcal{A}_1 }\{|
    \probP( f(\xneg,a_1)=1 \mid a_1=j )-
    \probP( f(\xneg,a_1)=1 )
|\} \,,\label{eqn,grp1,ext}
\end{equation}
and another meticulous formula inspired by DP \citep{jiang2020wasserstein} is  
\begin{equation}
\topequation%
\mathrm{DP}^\text{alt}(f) =\textstyle 
\max_{j,k\in\mathcal{A}_1, k\neq j}\{|
    \probP( f(\xneg,a_1)=1 \mid a_1=j )-
    \probP( f(\xneg,a_1)=1 \mid a_1=k )
|\} \,,\label{eqn,grp1,alt}
\end{equation}
as well as their corresponding average forms 
\begin{small}
\begin{align}
\topequation
\mathrm{DP}^\text{ext(avg)}(f) &=
\textstyle \frac{1}{n_{a_1}}
\sum_{ j\in\mathcal{A}_1 }\{|
    \probP( f(\xneg,a_1)=1 \mid a_1=j )-
    \probP( f(\xneg,a_1)=1 )
|\} \,,\label{eqn,grp1,ext,avg}\\
\mathrm{DP}^\text{alt(avg)}(f) &=\textstyle
\frac{2}{ n_{a_1}\!(n_{a_1}\!-1) } 
\sum_{j=1}^{n_{a_1}\!-\!1} 
\sum_{k=j\!+\!1}^{n_{a_1}} \{|
    \probP( f(\xneg,a_1)\!=\!1 | a_1\!=\!j )-
    \probP( f(\xneg,a_1)\!=\!1 | a_1\!=\!k )
|\} \,.\label{eqn,grp1,alt,avg}
\end{align}%
\end{small}%
Although these attempts capture finer distinctions, they essentially still rely on treating intersectional attributes as a single discrete variable. As a result, intersectionality remains only superficially addressed.

\begin{sidewaystable}
\centering
\caption{%
Summary of existing fairness measures. 
Note that `\#sen-att' denotes the number of sensitive attributes (\ie{} $n_a$). 
}\label{tab:fair,summary}
\renewcommand\tabcolsep{1.4pt}%
\begin{tabular*}{\textheight}{@{\extracolsep\fill}r|ccc|ccc|cc}
\toprule 
\multirow{2}{*}{\bf Name of {\em measure} \hspace{19mm}~} & 
\multirow{2}{*}{\textbf{Fairness type}\footnotemark[1]} & \multicolumn{2}{l|}{\bf Meaning} & \multicolumn{3}{l|}{\textbf{Applicable situation(s)} in definition} & 
\multicolumn{2}{l}{\textbf{Non-binary handling}\footnotemark[4]} \\
\cline{3-9} & & 
quant.\footnotemark[2] & fairer & \#label ($n_c$) & \#sen-att ($n_a$)\footnotemark[3] & \#values per $\mathcal{A}_i$ & $n_{a_i}\!>\!2$ & $n_a>1$ \\
\midrule
\tabincell{c}{Demographic parity \citep{gajane2017formalizing,jiang2020wasserstein} (\aka{}\\ statistical parity \citep{dwork2012fairness,chouldechova2017fair})} 
& *, group- & yes & lower value & binary & singular & bi-valued & yes & \reqdesign \\
Disparate impact /80\% rule \citep{feldman2015certifying,zafar2017fairness2} 
& *, group- & yes & \emph{larger} value & binary & singular & bi-valued & yes & \reqdesign \\
Disparate treatment \citep{zafar2017fairness2} 
& *, group- & \poss & lower value & binary & singular & 
bi- (multi- allowed) & yes & \reqdesign \\
Conditional statistical parity \citep{corbett2017algorithmic}
& *, group- & \poss & lower value & binary & singular & multi-valued & --- & \reqdesign \\
Bounded group loss \citep{agarwal2019fair} & *, group- & \poss & lower value & binary & singular & multi-valued & --- & \reqdesign \\
Strategic minimax fairness \citep{diana2024minimax} 
& *, group- & no & --- & bi-/multi- & singular & multi-valued & --- & \reqdesign \\
Equalised odds \citep{hardt2016equality,haas2019price} 
& *, group- & yes & lower value & binary & singular & bi-valued & yes & \reqdesign \\
Equality of opportunity \citep{hardt2016equality,gajane2017formalizing,haas2019price} 
& *, group- & yes & lower value & binary & singular & bi-valued & yes & \reqdesign \\
Predictive equality \citep{corbett2017algorithmic} 
& *, group- & \poss & lower value & binary & singular & multi-valued & --- & \reqdesign \\
$\gamma$-subgroup fairness \citep{kearns2018preventing,kearns2019empirical} 
& *, group- & yes & lower value & binary & singular & bi-valued & yes & \reqdesign \\
Predictive parity \citep{chouldechova2017fair,verma2018fairness} 
& *, group- & yes & lower value & binary & singular & bi-valued & yes & \reqdesign \\
Lipschitz condition \citep{dwork2012fairness,gajane2017formalizing} 
& *, individual- & no & --- & binary & singular & bi-valued & yes & \reqdesign \\
\tabincell{c}{General entropy indices \citep{speicher2018unified} (and \\the Theil index \citep{haas2019price})} 
& *, individual- & yes & lower value & binary & singular & multi-valued & --- & \reqdesign \\
Counterfactual fairness \citep{kusner2017counterfactual,gajane2017formalizing} 
& *, individual- & no & --- & binary & allows plural & multi- allowed & yes & \reqdesign \\
Proxy discrimination    \citep{kilbertus2017avoiding} 
& *, individual- & no & --- & binary & singular & multi- allowed & yes & \reqdesign \\
Discriminative risk \citep{bian2023increasing_alt} 
& *, \footnotemark[5] & yes & lower value & bi-/multi- & allows plural & multi-valued & --- & --- \\
Harmonic fairness via manifold 
& *, \footnotemark[5] & yes & lower value & bi-/multi- & allows plural & multi-valued & --- & --- \\
Multiaccuracy \citep{kim2019multiaccuracy} 
& *, group- & \poss & lower value & binary & singular & multi-valued & --- & \reqdesign \\
Differentially fair \citep{foulds2020intersectional} 
& *, group- & \poss & --- & binary & allows plural & bi- (multi- allowed) & yes & \reqdesign \\
\tabincell{c}{Group benefit ratio and worst-\\case min-max ratio \citep{ghosh2021characterizing}} 
& *, group- & yes & \emph{larger} value & binary & allows plural & bi- (multi-allowed) & yes & \reqdesign \\
Feature-apriori fairness   \citep{grgic2016case,grgic2018beyond} 
& procedural\,\footnotemark[6] & yes & --- & binary & --- & --- & yes & yes \\
Feature-accuracy fairness  \citep{grgic2016case,grgic2018beyond} 
& procedural\,\footnotemark[6] & yes & --- & binary & --- & --- & yes & yes \\
Feature-disparity fairness \citep{grgic2016case,grgic2018beyond} 
& procedural\,\footnotemark[6] & yes & --- & binary & --- & --- & yes & yes \\
FAE-based procedural fairness \citep{wang2024procedural} 
& procedural & yes & lower value & binary & singular & bi-valued & yes & \reqdesign \\
\bottomrule
\end{tabular*}
\footnotetext[1]{Mark * indicates the corresponding fairness is one of the \emph{distributive} fairness measures.}
\footnotetext[2]{Whether the corresponding fairness is a \emph{quantitative} measure. Note that many (including DP, EO, EOpp, and PP) may not be quantitative by definition, but remain a possibility to be one, indicated by `\poss'. DP, EO, EOpp, and PP are indicated as `yes' as there are widely used forms for them.}%
\footnotetext[3]{Whether it applies to intersectional SAs, in other words, `singular' and `plural' mean it can handle one \senatt{} only and more than one \senatt, respectively. }
\footnotetext[4]{Whether it can handle non-binary cases or whether it is possible to be extended. Note that `\reqdesign' means essentially it requires intersectional \sapl{} to be handled as one super discrete-valued (or multi-valued) \senatt.}
\footnotetext[5]{Discriminative risk (DR) is viewed as individual fairness primarily but can reflect group-level fairness as well; Harmonic fairness via manifold (HFM)~\citep{bian2024does,bian2024approximating} is based on distances between individuals and distances between sets, and therefore can reflect both individual- and group-level fairness.}
\footnotetext[6]{All three initial procedural fairness measures \citep{grgic2016case,grgic2018beyond} depend critically on member judgements regarding whether the use of particular features in decision-making is discriminatory. 
These member judgements may change over time, resulting in unstable outcomes and unpredictable computational demands as systems are recalibrated repeatedly. }
\end{sidewaystable}

\paragraph{Separation-based measures of fairness}
Equality of opportunity (EOpp) \citep{hardt2016equality,gajane2017formalizing,haas2019price} defined as
\begin{equation}
\topequation
\mathrm{EOpp}(f) =|
    \probP( f(\xneg,a_1)=1 \mid a_1=0, y=1 )-
    \probP( f(\xneg,a_1)=1 \mid a_1=1, y=1 )
| \,,\label{eqn,grp2}
\end{equation}
as well as relevant measures (such as equalised odds (EO) \citep{hardt2016equality,haas2019price} defined as
\begin{equation}
\topequation
\begin{split}
\mathrm{EO}= \tfrac{1}{2}[%
    &
    | \probP( f(\xneg,a_1)=1 \mid a_1=0, y=0 ) 
    - \probP( f(\xneg,a_1)=1 \mid a_1=1, y=0 ) | \\+&
    | \probP( f(\xneg,a_1)=1 \mid a_1=0, y=1 )
    - \probP( f(\xneg,a_1)=1 \mid a_1=1, y=1 ) |
]\,,\nonumber
\end{split}
\end{equation}
predictive equality, and $\gamma$-subgroup fairness), instantiates the separation criterion. They require that all groups experience the same false negative rate, which is associated with denied opportunities when acceptance is desired. 
Like DP, EOpp generalises imperfectly to multi-valued or intersectional settings, such as
\begin{subequations}
\topequation
\begin{align}
    \mathrm{EOpp}'(f) &=|
        \probP( f(\xneg,a_1)=1 \mid a_1\neq 1,y=1 )-
        \probP( f(\xneg,a_1)=1 \mid a_1=1, y=1 )
    | \,,\nonumber\\
    \mathrm{EOpp}^\text{ext}(f) &=
    \textstyle \max_{j\in\mathcal{A}_1}\{|
        \probP( f(\xneg,a_1)=1 \mid a_1=j, y=1 )-
        \probP( f(\xneg,a_1)=1 \mid y=1 )
    |\} \,,\nonumber\\
    \mathrm{EOpp}^\text{alt}(f) &= \textstyle
    \max_{j,k\in\mathcal{A}_1, k\neq j}\{|
        \probP( f(\xneg,a_1)=1 \mid a_1=j, y=1 )-
        \probP( f(\xneg,a_1)=1 \mid a_1=k, y=1 )
    |\} \,,\nonumber
\end{align}
\end{subequations}
as well as $\mathrm{EOpp}^\text{ext(avg)}$ and $\mathrm{EOpp}^\text{alt(avg)}$ analogous with Eq.~\eqref{eqn,grp1,ext,avg} to \eqref{eqn,grp1,alt,avg}, 
and binarisation and its extended formulas remain constrained.

\paragraph{Sufficiency-based measure of fairness}
Predictive parity (PP) \citep{chouldechova2017fair,verma2018fairness}, defined as 
\begin{equation}
\topequation
\mathrm{PP}(f) =|
    \probP( y=1\mid a_1=0, f(\xneg,a_1)=1 )-
    \probP( y=1\mid a_1=1, f(\xneg,a_1)=1 )
| \,,\label{eqn,grp3}
\end{equation}
is closely aligned with calibration and satisfies the sufficiency criterion. 
Extensions like
\begin{subequations}
\topequation
\begin{align}
    \mathrm{PP}'(f) &=|
        \probP( y=1\mid a_1\neq 1, f(\xneg,a_1)=1 )-
        \probP( y=1\mid a_1=1, f(\xneg,a_1)=1 )
    | \,,\nonumber\\
    \mathrm{PP}^\text{ext}(f) &=
    \textstyle \max_{j\in\mathcal{A}_1}\{|
        \probP( y=1\mid a_1=j, f(\xneg,a_1)=1 )-
        \probP( y=1\mid f(\xneg,a_1)=1 )
    |\} \,,\nonumber\\
    \mathrm{PP}^\text{alt}(f) &=\textstyle
    \max_{j,k\in\mathcal{A}_1, k\neq j}\{|
        \probP( y=1\mid a_1=j, f(\xneg,a_1)=1 )-
        \probP( y=1\mid a_1=k, f(\xneg,a_1)=1 )
    |\} \,,\nonumber
\end{align}
\end{subequations}
as well as $\mathrm{PP}^\text{ext(avg)}$ and $\mathrm{PP}^\text{alt(avg)}$ analogous with Eq.~\eqref{eqn,grp1,ext,avg} to \eqref{eqn,grp1,alt,avg}, 
again mirror those of DP and EOpp, allowing application to multi-valued attributes but inheriting the same limitations concerning intersectionality.

\paragraph{Why existing measures remain insufficient}
Taken together, \emph{group fairness focuses on statistical/demographic equality among groups defined by sensitive attributes} (with more definitions in Appendix~\ref{subsec:lit1}), and these measures offer useful but narrow perspectives on fairness. They often: 
(1) reduce multi-dimensional or intersectional identities to single discrete variables;
(2) conflate statistical parity with substantive fairness, potentially producing superficially ``fair'' yet socially misaligned outcomes; and 
(3) rely on simplified assumptions like binarisation that underestimate real disparities, or traversal that incurs computational costs. 
Therefore, while indispensable as baselines, these commonly used group fairness measures following the statistical criteria remain inadequate for advancing state-of-the-art (SOTA) fairness research. To address fairness in modern ML, it needs to go beyond them to frameworks that take structural inequities, intersectionality, and causal reasoning into account.

In contrast, \emph{individual fairness follows a principle that ``similar individuals should be evaluated or treated similarly.''} 
The Lipschitz condition \citep{dwork2012fairness} is primarily viewed as individual fairness, yet not a quantitative measure. 
Another famous example of individual fairness is counterfactual fairness (CFF) \citep{kusner2017counterfactual}, of which the analysis heavily relies on causal models to proceed with, meaning that learning algorithms other than causal graphs do not apply. 
Moreover, despite this apparent formalism, CFF is still typically constrained to a single \senatt. When intersectional or multiple attributes are involved, they must first be preprocessed to degenerate into one ``super'' attribute before the analysis, which not only introduces a higher computational burden but also risks obscuring the nuanced ways in which intersectional identities interact with structural biases. 
Except them, general entropy indices (GEI) \citep{speicher2018unified} and the Theil index (Theil) \citep{haas2019price} can serve as quantitative individual fairness measures, as well as two other fairness measures that can assess fairness from both individual and group aspects, that is, discriminative risk (DR) \citep{bian2023increasing_alt} and harmonic fairness via manifold (HFM) \citep{bian2024does,bian2024approximating}. 
It is worth noting that, in substance, most of the existing fairness in Table~\ref{tab:fair,summary} relies on degeneration into one discrete \senatt{} to handle intersectional attributes, except DR and HFM \citep{bian2024approximating}. 
The reason why DR and HFM differ from them is that DR does not need to divide data into (sub)groups by membership to compute, and that HFM tackles each \senatt{} separately and then puts them together, taking an analogous approach like the divide-and-conquer strategy.
More details about these individual fairness definitions and measures are given in Appendix~\ref{subsec:lit2} to \ref{subsec:lit3}.

\subsection{Empirical illustrations on the insufficiency of existing fairness measures}

Table~\ref{tab:fair,summary} highlights that most quantitative fairness measures are designed with a single, typically binary, \senatt{} in mind. While several extensions allow for applications to multi-valued attributes, our experiments reveal substantive risks in such adaptations, underscoring the need for fairness measures that are carefully constructed to address non-binary and intersectional attributes.

\begin{figure}[tb]\centering
\begin{minipage}{\textwidth}\centering
\subfloat[]{\includegraphics[height=3.3cm]{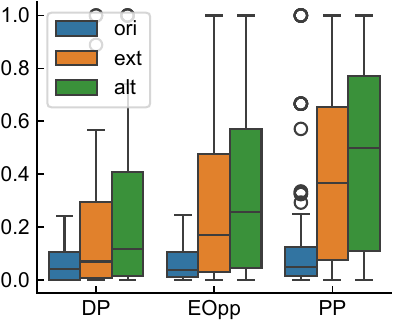}}
\hspace{1mm}
\subfloat[]{\includegraphics[height=3.3cm]{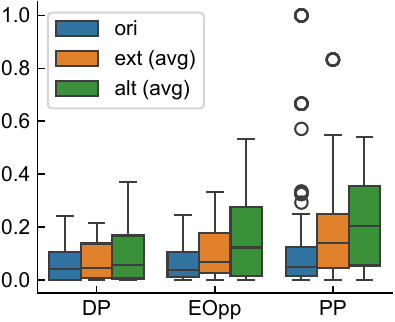}}
\hspace{1mm}
\subfloat[]{\includegraphics[height=3.3cm]{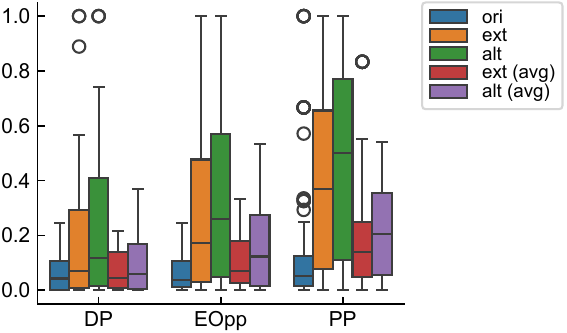}}
\vspace{-1mm}\caption{\small 
Comparison of three commonly used group fairness measures and their extensions, on Income, Compas PPR, and Compas PPVR datasets. 
(a) Comparison between their original definitions and the first two extension forms, analogously to Eq.~\eqref{eqn,grp1}, \eqref{eqn,grp1,ext}, and \eqref{eqn,grp1,alt}; note that \eqref{eqn,grp1,bin} in binarisation is equivalent to \eqref{eqn,grp1}. 
(b) Comparison between their original definitions and the last two extension forms, analogously to Eq.~\eqref{eqn,grp1,bin}, \eqref{eqn,grp1,ext,avg}, and \eqref{eqn,grp1,alt,avg}. 
(c) Comparison between their original definitions and all four extension formulas. 
}\label{fig:eqn,grp}
\vspace{-4mm}
\end{minipage}
\end{figure}
\begin{figure}[tb]\centering
\begin{minipage}{\textwidth}\centering
\subfloat[]{\includegraphics[height=3.2cm]{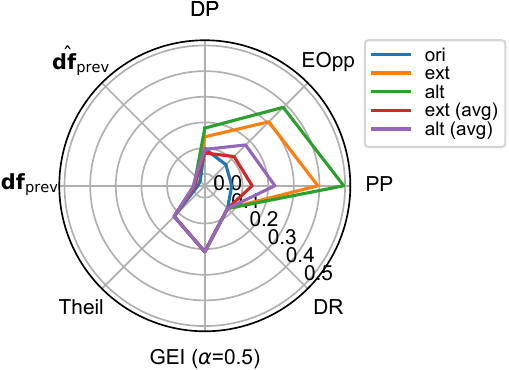}}
\subfloat[]{\includegraphics[height=3.2cm]{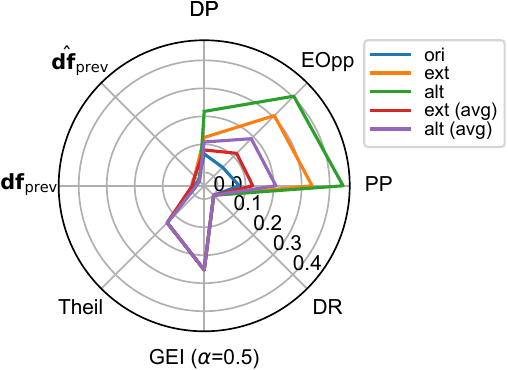}}
\subfloat[]{\includegraphics[height=3.2cm]{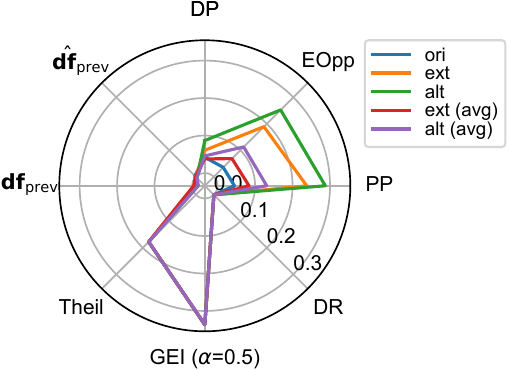}}
\vspace{-3mm}\\
\subfloat[]{\includegraphics[height=3.2cm]{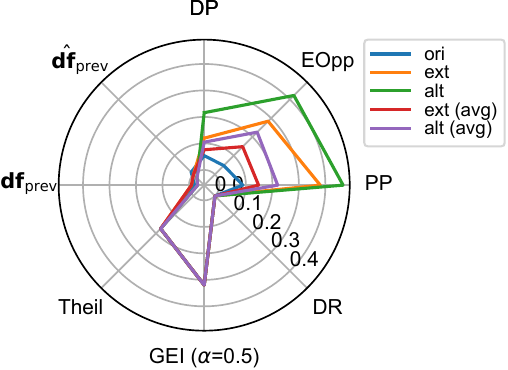}}
\subfloat[]{\includegraphics[height=3.2cm]{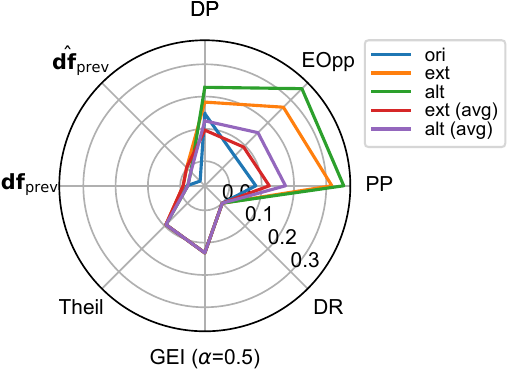}}
\subfloat[]{\includegraphics[height=3.2cm]{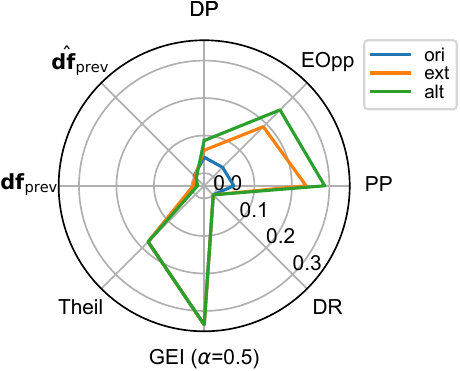}}
\hspace{1mm}
\vspace{-1mm}\caption{\small
Comparison of fairness measures between their original definitions and their corresponding extension formulas, on the Income dataset.
(a--e) Using bagging, AdaBoost, LightGBM, AdaFair (trained using the first \senatt), and AdaFair (trained using the second \senatt), respectively; (f) Using LightGBM.
Note that the previous HFM in one bi-valued \senatt{} and its approximated value are indicated as $\dfprev$ and $\dfpreva$, respectively, as the original definition of HFM; the maximum HFM for one multi-valued \senatt{} and its approximated value are used as their extension and `ext(avg)' forms; and the average HFM for one multi-valued \senatt{} and its approximated value are used as their alternative and `alt(avg)' forms. 
}\label{fig:radar,adult}
\vspace{-3mm}
\end{minipage}
%
%
%
%
%
\begin{minipage}{\textwidth}
\centering%
\subfloat[]{\label{subfig:tim,a}
\includegraphics[height=3.2cm]{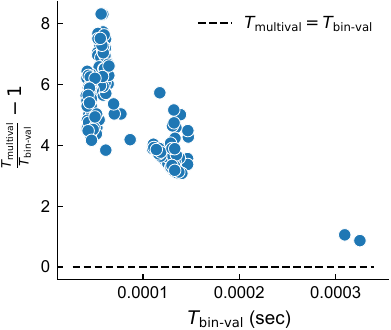}}
\subfloat[]{\label{subfig:tim,na}%
\includegraphics[height=3.2cm]{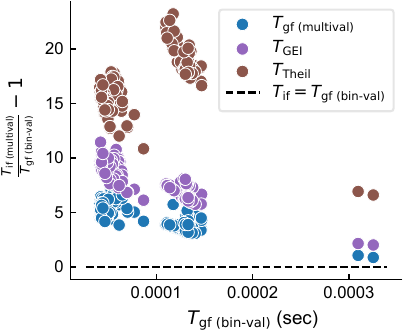}}
\hspace{.5mm}
\subfloat[]{\label{subfig:tim,b}
\includegraphics[height=3.2cm]{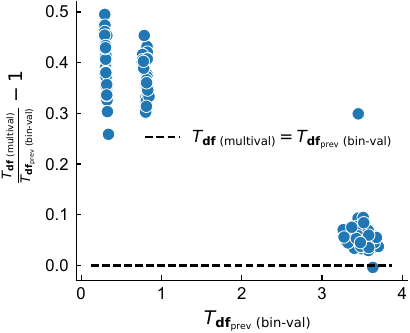}}
\hspace{.5mm}
\subfloat[]{\label{subfig:tim,c}
\includegraphics[height=3.2cm]{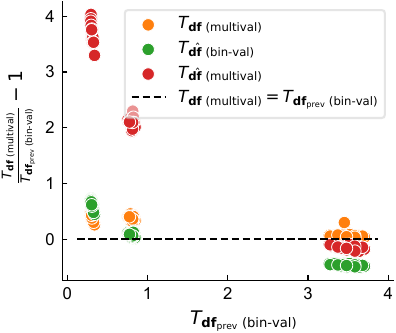}}
\vspace{-3mm}\\
\subfloat[]{\label{subfig:tim,d}
\includegraphics[height=3.2cm]{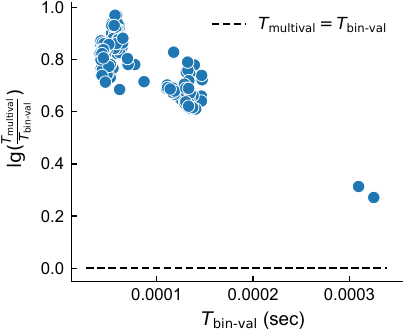}}
\subfloat[]{\label{subfig:tim,nb}%
\includegraphics[height=3.2cm]{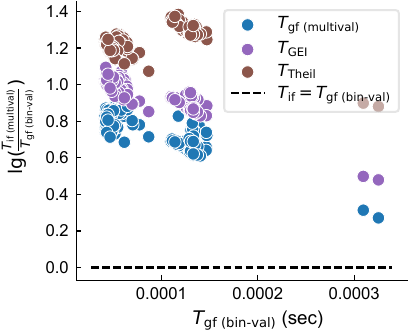}}
\subfloat[]{\label{subfig:tim,e}
\includegraphics[height=3.2cm]{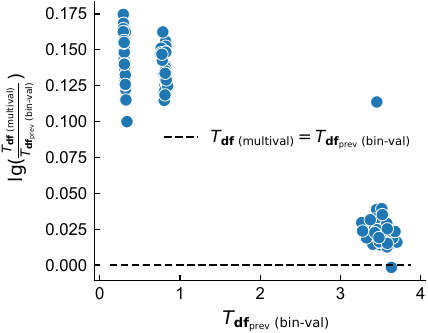}}
\subfloat[]{\label{subfig:tim,f}
\includegraphics[height=3.2cm]{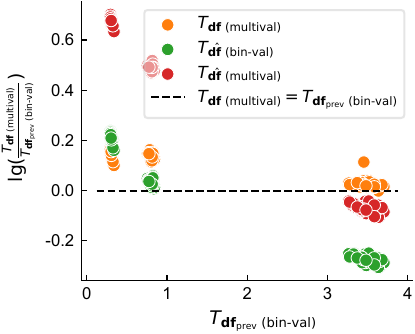}}
\vspace{-1mm}\caption{\small%
Time cost comparison at different scales, for Income, Compas PPR, and Compas PPVR datasets. Note that the dashed line indicates the same time cost for computation, separating larger (above) and smaller (below) computation costs.
(a) and (e) Comparison of three commonly used group fairness measures and their extension formulas; 
(b) and (f) Comparison between group fairness and individual fairness (GEI and Theil); 
(c--d) and (g--h) Time cost comparison of HFM for binary-value and multi-value cases, where (c) and (g) are comparisons of direct computation only, and (d) and (h) are comparisons including the results using approximation algorithms \citep{bian2024does,bian2024approximating}.
}\label{fig:tim}
\end{minipage}
\end{figure}

\paragraph{Binarisation underestimates discrimination}
As shown in Figure~\ref{fig:eqn,grp}, binarising a multi-valued \senatt{} (\eg{} Eq.~\eqref{eqn,grp1,bin} and analogues) systematically yields smaller values compared with extensions such as Eq.~\eqref{eqn,grp1,ext} and \eqref{eqn,grp1,alt}. Even average-based forms in Eq.~\eqref{eqn,grp1,ext,avg} and \eqref{eqn,grp1,alt,avg} often detect stronger disparities than binarisation. This pattern recurs across datasets regardless of the learning algorithms in use (shown in Figure~\ref{fig:radar,adult}, and Figures \ref{fig:radar,ppr} to \ref{fig:radar,ppvr} of Appendix~\ref{sec:appx}), and suggests that binarisation oversimplifies the structure of disadvantage and can systematically underestimate discrimination. In contrast, individual fairness measures that naturally accommodate multi-valued attributes (such as GEI, Theil, DR, and HFM) remain stable across these settings. Given that even modest underestimations of bias can accumulate and amplify improper prejudice in human-AI interactions~\citep{glickman2024human,vlasceanu2022propagation}, accurate measurement is essential for responsible deployment of ML systems.

\paragraph{Traversal-based generalisation incurs computational cost}%
Extending group fairness measures (\ie{} DP, EOpp, and PP) to multi-valued \sapl{} is not only conceptually challenging but also computationally demanding. 

Figures~\ref{fig:tim}\subref{subfig:tim,a} and \ref{fig:tim}\subref{subfig:tim,d} show that formulas such as Eq.~\eqref{eqn,grp1,ext} and \eqref{eqn,grp1,alt} require almost an order of magnitude more time than their binarised counterparts like Eq.~\eqref{eqn,grp1,bin} in such multi-valued scenarios, and note that this is only the results for one 5- or 6-valued \senatt. 
As for individual fairness measures that can handle one multi-valued \senatt{} (\eg{} GEI and Theil), they have an even heavier computational burden, doubling to quadrupling that of the extensions of group fairness, presented in Figures~\ref{fig:tim}\subref{subfig:tim,na} and \ref{fig:tim}\subref{subfig:tim,nb}. 
Similar observations that handling multi-valued cases brings about more computational costs recur in the remaining sub-figures in Figure~\ref{fig:tim}: 
Even for HFM, its maximum and average versions~\citep{bian2024approximating} that can handle one multi-valued \senatt{} increase runtime by up to 1.5$\times$ compared with its previous bi-valued version~\citep{bian2024does}, presented in Figures~\ref{fig:tim}\subref{subfig:tim,b} and \ref{fig:tim}\subref{subfig:tim,e}; Approximation strategies \citep{bian2024does,bian2024approximating} offer partial relief, presented in Figures~\ref{fig:tim}\subref{subfig:tim,c} and \ref{fig:tim}\subref{subfig:tim,f}, yet still lag behind simpler formulations. 

Furthermore, it is obvious that degenerating intersectional attributes into one ``super'' discrete \senatt{} through preprocessing is not an efficient way to handle them. 
For instance, consider data with two \sapl{} $\mathcal{A}= \mathcal{A}_1\!\times\!\mathcal{A}_2 = \mathbb{Z}^{n_{a_1}}\!\times\!\mathbb{Z}^{n_{a_2}}$, where $n_{a_1}, n_{a_2} \!\geqslant\! 2$. 
After preprocessing, the combined single \senatt{} would be $\mathcal{A}'= \mathbb{Z}^{n_{a_1}\!\times n_{a_2}}$. 
This approach may be practical when both $n_{a_1}$ and $n_{a_2}$ are small enough, yet the computational cost increases exponentially as these values grow (\eg{} if $n_{a_1}=2$ and $n_{a_2}$ changes from $2$ to $6$, $\mathcal{A}'$ transitions from $\mathbb{Z}^4$ to $\mathbb{Z}^{12}$). Similarly, if the number of \sapl{} itself increases, the computational burden also rises (\eg{} $\mathcal{A}'$ could shift from $\mathbb{Z}^{2\times 6} \!=\! \mathbb{Z}^{12}$ to $\mathbb{Z}^{2\times 6\times 3} \!=\! \mathbb{Z}^{36}$). 
The larger the number of values for each \senatt, the longer it takes to compute. 
In such cases, the computational cost may be reduced if one fairness measure can handle multiple \sapl{} directly. This approach can be viewed as decomposing the original complex problem into smaller and more manageable sub-problems, akin to the divide-and-conquer strategy.

These results reveal that efficient fairness assessment for multi-valued and even intersectional attributes remains an open practical challenge, motivating the development of more computationally tractable approaches.

\begin{figure}[tb]
\begin{minipage}{\textwidth}
\centering%
\subfloat[]{\label{subfig:corr,a}
\includegraphics[height=3.3cm]{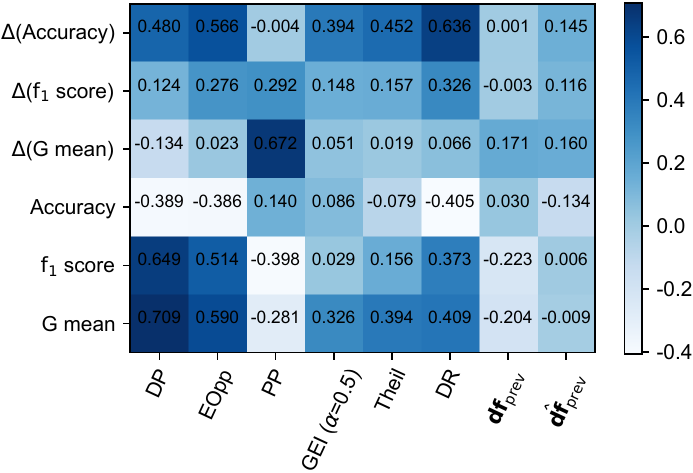}}
\subfloat[]{\label{subfig:corr,b}
\includegraphics[height=3.3cm]{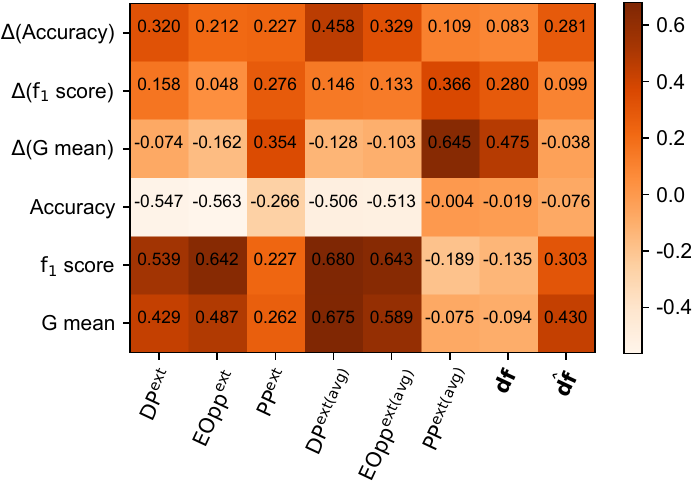}}
\subfloat[]{\label{subfig:corr,c}
\includegraphics[height=3.3cm]{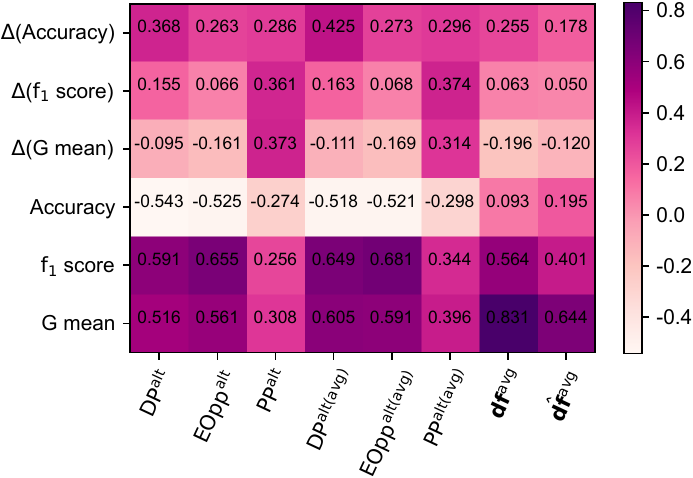}}
\vspace{-1mm}\caption{\small%
The correlation between the performance change due to sensitive attributes only and different fairness measures, using Pearson's correlation coefficients.
(a) Using the original definitions of group fairness measures, equivalent to \eqref{eqn,grp1,bin} and its analogues, and the previous HFM \citep{bian2024does} and its approximated results, denoted by $\hat{\mathbf{df}}_\text{prev}$; 
(b) Using \eqref{eqn,grp1,ext}, \eqref{eqn,grp1,ext,avg}, and their analogues, as well as the maximum HFM \citep{bian2024approximating} and its approximation, denoted by $\hat{\mathbf{df}}$; 
(c) Using \eqref{eqn,grp1,alt}, \eqref{eqn,grp1,alt,avg}, and their analogues, as well as the average HFM \citep{bian2024approximating} and its approximation, denoted by $\hat{\mathbf{df}}^\text{avg}$. 
}\label{fig:corr}
\vspace{-4mm}
\end{minipage}
\begin{minipage}{\textwidth}
\centering
\subfloat[]{\label{subfig,to,a}%
\includegraphics[height=3.3cm]{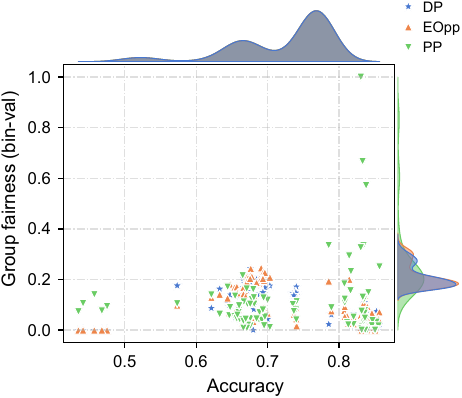}}\hspace{3mm}
\subfloat[]{\label{subfig,to,b}%
\includegraphics[height=3.3cm]{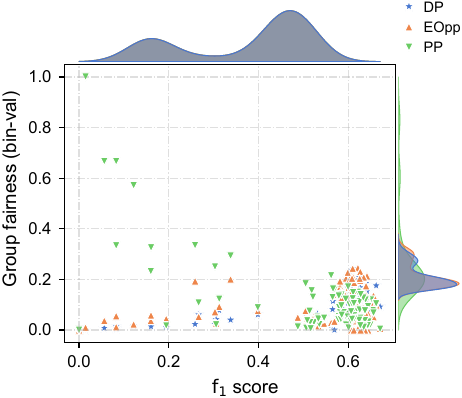}}\hspace{4mm}
\subfloat[]{\label{subfig,to,c}%
\includegraphics[height=3.3cm]{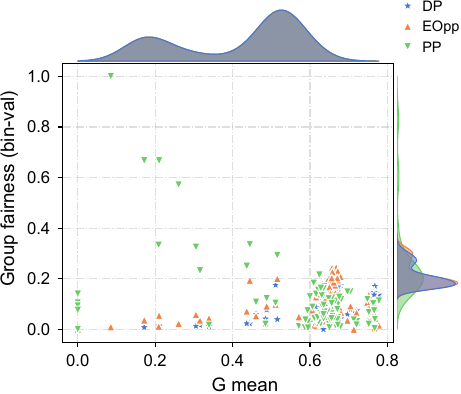}}
\vspace{-3mm}\\
\subfloat[]{\label{subfig,t/o,d}%
\includegraphics[height=3.3cm]{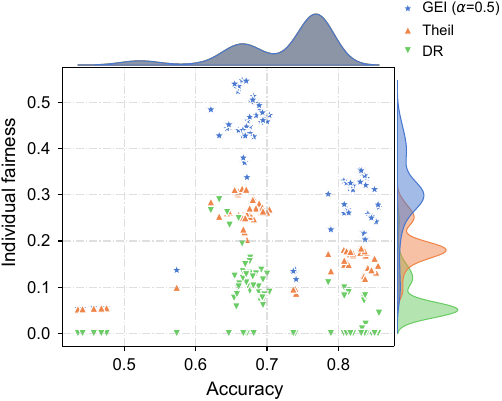}}\hspace{1mm}
\subfloat[]{\label{subfig,t/o,e}%
\includegraphics[height=3.3cm]{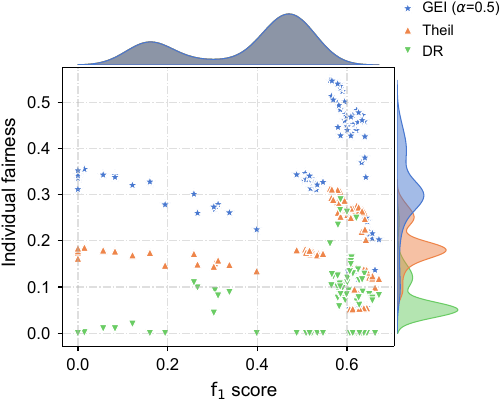}}\hspace{1mm}
\subfloat[]{\label{subfig,t/o,f}%
\includegraphics[height=3.3cm]{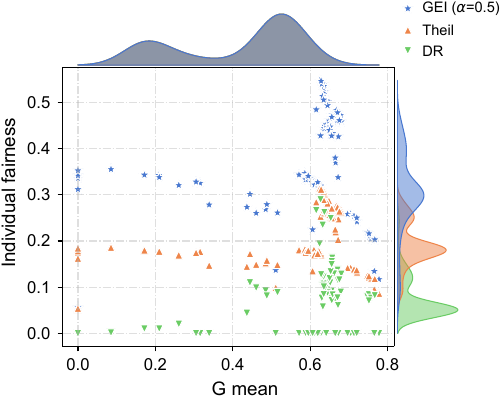}}
\vspace{-1mm}\caption{\small%
Scatter plot between performance (accuracy, $\mathrm{f}_1$ score, or geometric mean \citep{akosa2017predictive}, respectively) and fairness.
Note that on the $y$-axis, the smaller the better; on the $x$-axis, the larger the better. 
(a--c) Using three commonly used group fairness measures, equivalent to \eqref{eqn,grp1,bin} and its analogous formulas; 
(d--f) Using individual fairness measures. 
}\label{fig:trade-off}%
\end{minipage}
\end{figure}

\begin{figure}[tb]
\centering%
\subfloat[]{\label{subfig:rel,d}%
\includegraphics[height=3.4cm]{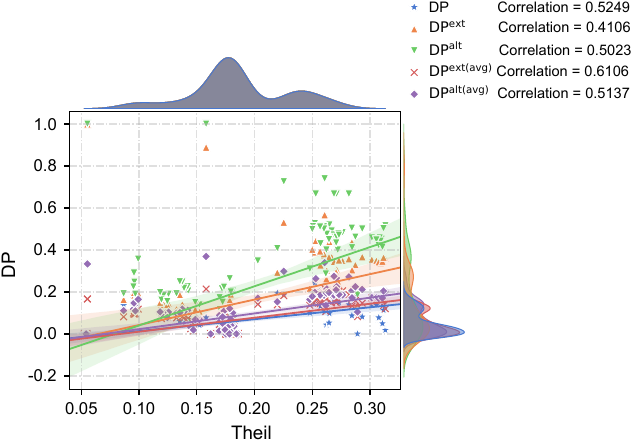}}
\subfloat[]{\label{subfig:rel,e}%
\includegraphics[height=3.4cm]{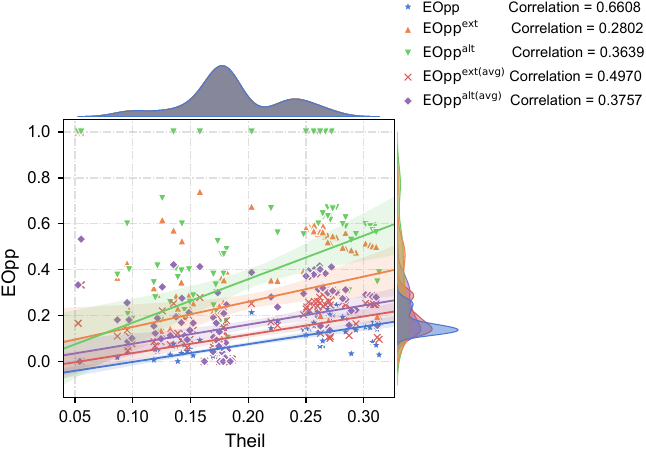}}
\subfloat[]{\label{subfig:rel,f}%
\includegraphics[height=3.4cm]{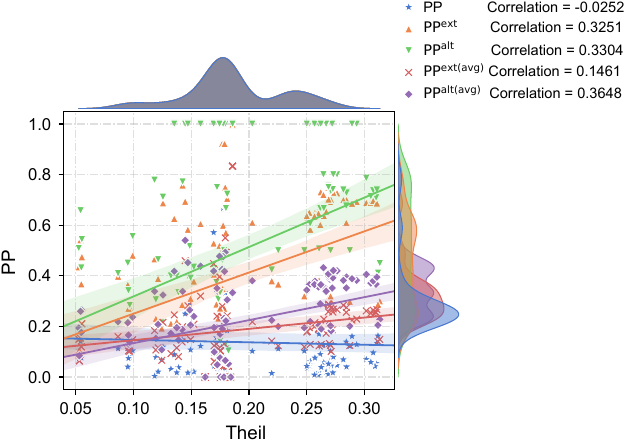}}
\vspace{-3mm}\\
\subfloat[]{\label{subfig:rel,g}%
\includegraphics[height=3.4cm]{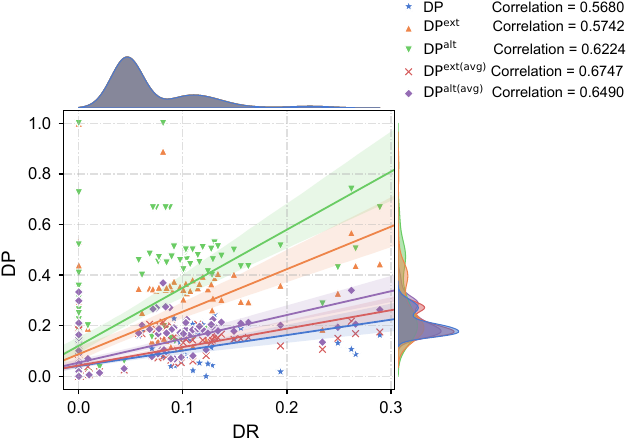}}
\subfloat[]{\label{subfig:rel,h}%
\includegraphics[height=3.4cm]{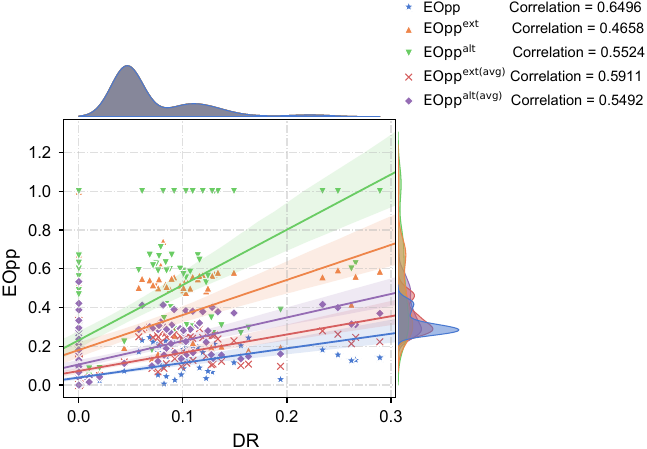}}
\subfloat[]{\label{subfig:rel,i}%
\includegraphics[height=3.4cm]{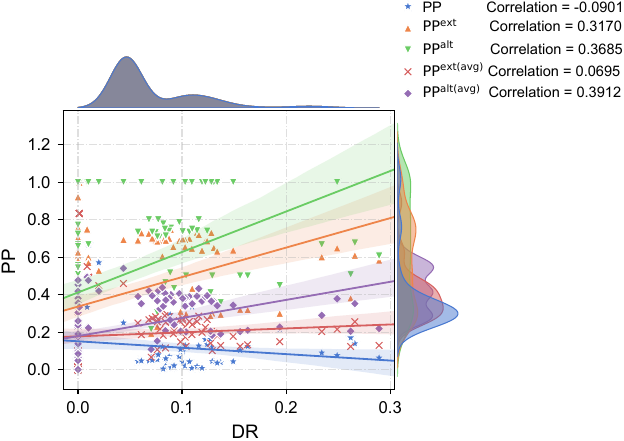}}
\vspace{-3mm}\\
\subfloat[]{\label{subfig:rel,cd1,a}%
\includegraphics[height=3.4cm]{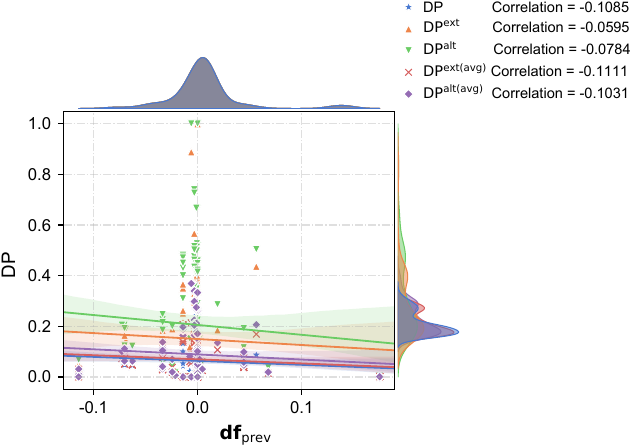}}
\subfloat[]{\label{subfig:rel,cd1,b}%
\includegraphics[height=3.4cm]{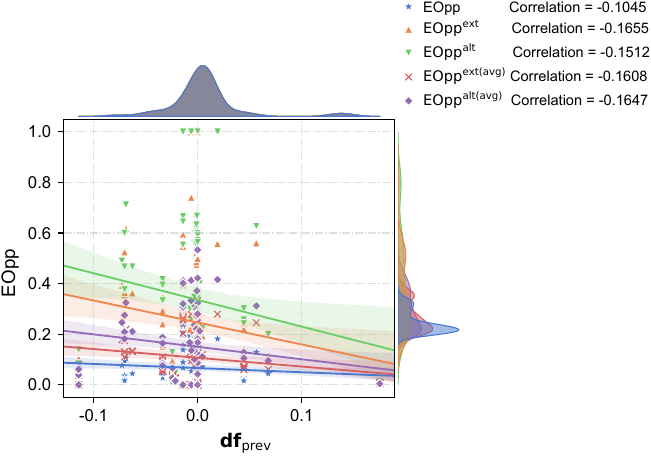}}
\subfloat[]{\label{subfig:rel,cd1,c}%
\includegraphics[height=3.4cm]{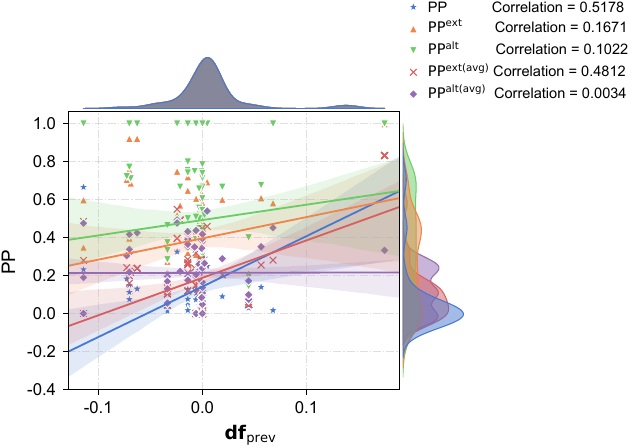}}
\vspace{-3mm}\\
\subfloat[]{\label{subfig:rel,cd1,d}%
\includegraphics[height=3.4cm]{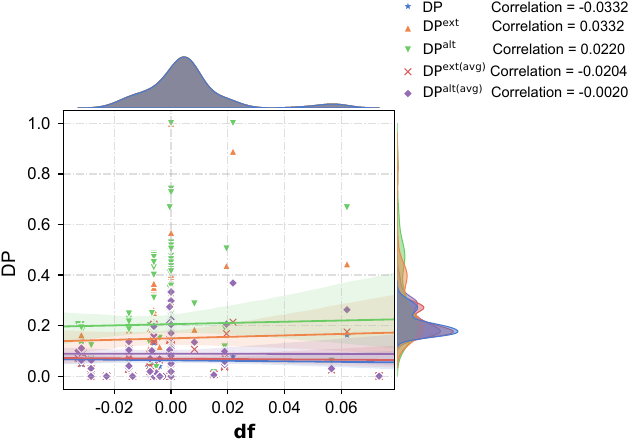}}
\subfloat[]{\label{subfig:rel,cd1,e}%
\includegraphics[height=3.4cm]{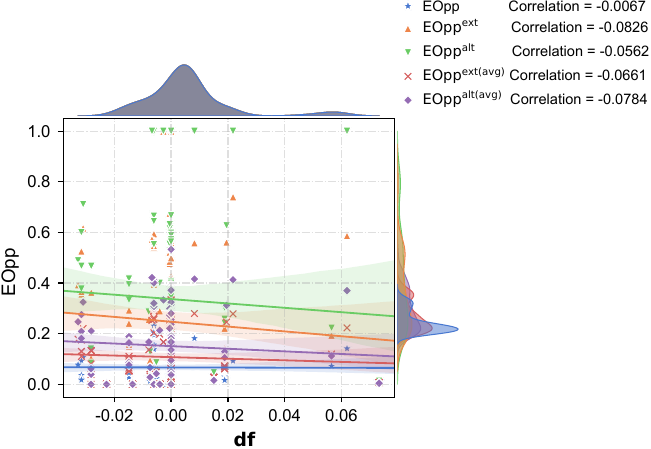}}
\subfloat[]{\label{subfig:rel,cd1,f}%
\includegraphics[height=3.4cm]{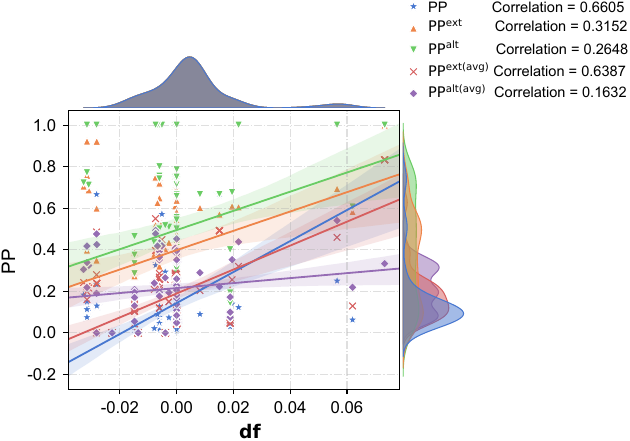}}
\vspace{-1mm}\caption{\small
Relation between individual fairness and group fairness (DP, EOpp, and PP), on the Income, Compas PPR, and Compas PPVR datasets. Note that on both $x$- and $y$- axes, the smaller the better. 
(a--c), (d--f), (g--i), and (j--l) Using Theil, DR, the previous HFM \citep{bian2024does}, and the maximum HFM \citep{bian2024approximating} as the individual fairness, respectively. 
}\label{fig:reln}%
\end{figure}

\paragraph{Extension formulas fail to capture discrimination robustly}
We further examine correlations between fairness measures and $\Delta$(performance), defined as performance changes when only \sapl{} are perturbed. Figure~\ref{fig:corr} shows that extended group fairness measures typically exhibit weaker correlations with $\Delta$(performance) than their original or binarised versions. By contrast, DR aligns closely with $\Delta$(Accuracy), indicating that it directly captures discrimination arising from changes in \sapl. Moreover, incompatibilities between fairness criteria persist~\citep{barocas2023fairness}, meaning that satisfying one measure does not preclude unfairness by another. Together, these findings indicate that current extensions to multi-valued settings are insufficiently sensitive, and that more robustly formulated metrics are needed to capture discrimination faithfully across diverse attribute structures.

\subsection{Empirical illustrations on incompatibility views}

\paragraph{Accuracy and fairness are not strictly incompatible}
A prevailing view in the literature describes fairness and accuracy as fundamentally at odds, with improvements in one often assumed to degrade the other \citep{berk2021fairness}. Our experiments shown in Figure~\ref{fig:trade-off} (and Figure \ref{fig:t/o,cont} of Appendix~\ref{sec:appx}) partly support this view: when models operate at low to moderate accuracy, gains in predictive performance can indeed exacerbate disparities. However, the relationship is more nuanced. At sufficiently high levels of accuracy, improvements can coincide with enhanced fairness, particularly in the case of individual fairness, demonstrated in Figures~\ref{fig:trade-off}\subref{subfig,t/o,d} to \ref{fig:trade-off}\subref{subfig,t/o,f}. For group fairness, this alignment is less obvious in the original formulations, but becomes clearer in their extended forms presented in Figures \ref{fig:t/o,cont}\subref{subfig,toc,a}, \ref{fig:t/o,cont}\subref{subfig,toc,d}, \ref{fig:t/o,cont}\subref{subfig,toc,g}, and \ref{fig:t/o,cont}\subref{subfig,toc,j}, as well as the remaining sub-figures of Figure~\ref{fig:t/o,cont}. These findings suggest that the long-assumed trade-off is not universal; instead, it points to opportunities for promoting fairness and accuracy jointly, especially by prioritising individual fairness in high-performance regimes.

\paragraph{Individual fairness and group fairness are not inherently incompatible}

In the literature, group fairness is frequently framed as incompatible with individual fairness \citep{hardt2016equality,pleiss2017fairness}: for example, satisfying group fairness still permits predicting qualified applicants in one demographic but random individuals in another. 
Yet the reverse may not be the case: Even intuitively, if individuals are all treated fairly, the fair treatment concerning their demographic is supposed to follow, meaning that the satisfaction of individual fairness should achieve group fairness as well. 
This intuition is empirically demonstrated in Figures~\ref{fig:reln}\subref{subfig:rel,d}, \ref{fig:reln}\subref{subfig:rel,e}, \ref{fig:reln}\subref{subfig:rel,g}, and \ref{fig:reln}\subref{subfig:rel,h}: 
The worsen individual fairness (Theil and DR) signifies the same worsen tendency of group fairness (DP and EOpp), as well as analogously when individual fairness gets better. 
The same trend of change is also evidenced by their correlations: Figures~\ref{fig:reln}\subref{subfig:rel,d} and \ref{fig:reln}\subref{subfig:rel,e} show that Theil is moderately correlated with DP, EOpp, and the average forms of DP's extensions; Figures~\ref{fig:reln}\subref{subfig:rel,g} and \ref{fig:reln}\subref{subfig:rel,h} show that DR is moderately and nearly highly correlated with DP, EOpp, and the average forms of their corresponding extensions. 

It is worth noting a misconception here: Some may exemplify hiring a candidate from the marginalised group(s) and claim that meeting individual fairness cannot guarantee group fairness. However, hiring a candidate from the marginalised group(s) itself does not necessarily mean individual fairness is satisfied, because it is possible that one views the hired candidate from the marginalised group(s) as less qualified than their competitors from the privileged group and gets hired mainly due to policy reasons, which is still discrimination. 
Moreover, some may argue that the distributional differences across groups may exist and can lead to aggregated disparities even under perfect individual fairness, yet ignoring historical reasons and misattributing those differences to \sapl{}, without being aware of the lack of purely clean or fair data, followed by another difficulty of mitigating discrimination in reality.

Besides, neither Theil nor DR shows a high correlation with PP in Figures~\ref{fig:reln}\subref{subfig:rel,f} and \ref{fig:reln}\subref{subfig:rel,i}, which also supports the incompatibility among the three statistical non-discrimination criteria; similar observation can also be found in the average HFM by Figures~\ref{fig:rel,contd1}\subref{subfig:rel,cd1,g} to \ref{fig:rel,contd1}\subref{subfig:rel,cd1,i}. 
The incompatibility among these criteria is also supported by Figures~\ref{fig:reln}\subref{subfig:rel,cd1,a} to \ref{fig:reln}\subref{subfig:rel,cd1,f}, where the previous and maximum HFM have moderate to nearly high correlation with PP, yet are almost irrelevant to DP and EOpp. 
Note that HFM, differing from all other fairness measures, can only capture the extra bias introduced in the learning procedure; the reason why the average HFM exhibits different correlations with group fairness may be that the average HFM among the three versions is not as extreme as its two counterparts using the maximal operator, therefore, more consistent with discrimination evaluated via predictions. 
As for GEI, it shows a moderate correlation with EOpp only, presented in Figure~\ref{fig:rel,contd1}\subref{subfig:rel,b}, and therefore has fewer values as a signal compared to Theil and DR.

\subsection{Worth-considering principles in the design of fairness metrics}

Our empirical analyses above uncover structural limitations in extending existing fairness measures to non-binary \sapl: 
(1) Binarisation oversimplifies and underestimates discrimination systematically for non-binary \sapl; 
(2) Traversal-based extensions suffer from substantial computational burden and often misalign with discrimination-sensitive performance shifts. 
These limitations highlight a pressing need for fairness measures that are both conceptually rigorous and practically scalable. 
Addressing this challenge requires a paradigm shift: instead of serving as a peripheral constraint in predictive modelling, fairness should be treated as a measurable property that evolves with a deeper understanding of its complexity and the societal impact of modern ML systems, alongside the increasing scale.

Looking forward, we argue that reasonable and viable fairness metrics in practice should follow a main line and be guided by the following design principles:
\begin{itemize}
\item \textbf{Conceptual coherence}: It ought to take societal considerations where discrimination originates from into account and minimise the incompatibility or contradictions across definitions, offering a more unified and interpretable view of fairness. 
\item \textbf{Modelling faithfulness}: It should reflect disparities as they manifest in real-world outcomes, aligning with performance shifts when \sapl{} are perturbed, and can capture disparities across multi-valued and intersectional attributes, avoiding reductive binarisation. 
\item \textbf{Computational tractability}: It is supposed to scale efficiently to realistic datasets and models without prohibitive cost, and remain robust under practical deployment conditions, where small biases risk accumulation and amplification in human-AI interactions.
\end{itemize}
By embedding these principles, we cover the field can move beyond incremental adaptations of binary-group measures and towards fairness metrics that more faithfully reflect the structural complexity of discrimination in modern ML systems, supporting more equitable, trustworthy, and socially aligned AI.

\section{Final remarks}

In this paper, we revisited and clarified several persistent misconceptions in the study of algorithmic fairness. By comprehensively analysing fairness measures that are widely used, we presented that most existing formulations remain bound to one single binary \senatt, and that their naive extensions to multi-valued or intersectional attributes can lead to severe underestimation of discrimination, inflated computational costs, and weak alignment with performance-sensitive disparities. Through empirical studies, we also demonstrated that fairness and accuracy, as well as individual and group perspectives, need not be inherently incompatible, thereby challenging conventional wisdom in the field. Together, these findings provide a clearer foundation for advancing fairness research beyond the restrictive assumptions that dominate much of the current discourse.

\begin{appendices}

\section{Related work: summarising existing fairness definitions and measures}
\label{sec:related}

In this section, we summarise existing definitions or measures of fairness, starting with a list of some standard notations that we use throughout this paper. 
Then we separately introduce distributive fairness (centred around the outcomes/predictions of the decision process) in Sections~\ref{subsec:lit1} to \ref{subsec:lit3} and procedural fairness in Section~\ref{subsec:lit4}, where the latter is based on the decision-making process rather than the outcomes.

\subsection{Preliminaries}
\label{subsec:lit0}
In this paper, we denote
\begin{itemize}
\setlength{\itemsep}{-.4mm}%
\item scalars by italic lowercase letters (\eg{} $x$),
\item vectors by bold lowercase letters (\eg{} $\mathbf{x}$),
\item matrices/sets by italic uppercase letters (\eg{} $X$),
\item random variables by serif uppercase letters (\eg{} $\mathsf{X}$),
\item real numbers (resp. integers, and positive integers) by $\mathbb{R}$ (resp. $\mathbb{Z}$, and $\mathbb{Z}_+$),
\item probability measure (resp. expectation, and variance of one random variable) by $\mathbb{P}(\cdot)$ (resp. $\mathbb{E}(\cdot)$, and $\mathbb{V}(\cdot)$),
\item hypothesis space (resp. models or predictors or learners or classifiers in this space) by $\mathcal{F}$ (resp. $f(\cdot)$).
\end{itemize}

Here, for one unknown distribution $\probD$ over $\mathcal{X\times A\times Y}= \mathbb{R}^{n_d}\times \mathbb{Z}^{n_a}\times \{0,1,...,n_c-1\}$ where $\mathcal{X\times A}$ and $\mathcal{Y}$ are, respectively, the feature/input space and label/output space. 
Note that $n_a, n_d, n_c\in\mathbb{Z}_+$ and they are the numbers of protected/sensitive attributes, unprotected/insensitive/remaining attributes, and labels, respectively. 
In general, vanilla ML tasks aim to optimise the expected predictive accuracy, that is, minimising the expected accuracy risk 
\begin{equation}
\topequation
    \err(f)= 1-\acc(f)=
    \mathbb{E}_{(\xneg,\xpos,y)\sim \probD}[
        \mathbb{I}( f(\xneg,\xpos) \neq y )
    ] \,.\label{eq:acc}
\end{equation}
Other commonly-used performance metrics include precision, recall/sensitivity (\aka{} true positive rate), specificity (\aka{} true negative rate), $\mathrm{f}_1$ score, area under curve (AUC), false positive rate, and false negative rate.

The tasks relevant to discrimination/bias mitigation are usually binary (that is, $\mathcal{Y}=\{0,1\}$ or $\{+1,-1\}$, and $n_c=2$); we will explicitly indicate it if the analysis also applies to multi-class classification (that is, $n_c\geqslant 3$). 
In these tasks, usually a predictor function $f:\mathcal{X\times\mathcal{A}} \mapsto\mathcal{Y}$ is expected to be learned. 
A scoring function $h: \mathcal{X\times\mathcal{A}} \mapsto[0,1]$ may be introduced in some literature, and in that case, 
\begin{equation}
    \topequation
    f(\xneg,\xpos)=1 \text{ \iff{} } 
    h(\xneg,\xpos) \geqslant \text{threshold}
    \,.\label{eq:fxh}
\end{equation}

It is worth noting that most existing fairness measures in the literature only work for one single sensitive attribute (\senatt), 
that is to say, $n_a=1$, and $\mathcal{A}=\mathcal{A}_1\subseteq\mathbb{Z}$ is a finite set, and it is usually with binary values (that is, $\mathcal{A}_1=\{0,1\}$ or $\{1,2\}$); we will explicitly indicate it if the corresponding assessment also applies to multiple \sapl{} (that is, $n_a\geqslant 2$). 
Throughout this paper, we keep using $a_1=1$ to represent the corresponding instance belonging to the privileged group and $a_1\neq 1$ that belonging to marginalised group(s), which is suitable for both $n_{a_1}\defineq |\mathcal{A}_1|=2$ and $n_{a_1}>2$ where $\mathcal{A}_1=\{1,2,...,n_{a_1}\}$.

Note that, in the following, we may abbreviate $\mathbb{E}_{(\xneg,\xpos,y)\sim \probD}[\cdot]$ and $\mathbb{P}_{(\xneg,\xpos,y)\sim \probD}(\cdot)$ as $\probE[\cdot]$ and $\probP(\cdot)$, respectively, for brevity when the context is unambiguous. 
We also use $i\in[n]$ to represent $i\in\{1,2,...,n\}$ for brevity.

\subsection{Group fairness}
\label{subsec:lit1}

\begin{definition}[Demographic parity \citep{gajane2017formalizing,jiang2020wasserstein}, \aka{} statistical parity \citep{dwork2012fairness,chouldechova2017fair}, equal acceptance rate \citep{zliobaite2015relation,verma2018fairness}, or benchmarking \citep{simoiu2017problem,verma2018fairness}]
\label{def:grp1,dp}%
\begin{equation}
\topequation
    \probP( f(\xneg,a_1)=1 \mid a_1=0 )=
    \probP( f(\xneg,a_1)=1 \mid a_1=1 )
\,,\label{eq:grp,dp}
\end{equation}
or in the form of a scoring function \citep{chouldechova2017fair}
\begin{equation}
\topequation
\probP( h(\xneg,a_1)> \thres \mid a_1=0 )=
\probP( h(\xneg,a_1)> \thres \mid a_1=1 ) 
\,. 
\end{equation}
It also has an essentially similar form called overall accuracy equality \citep{verma2018fairness},
\begin{equation}
\topequation
\probP( f(\xneg,a_1)=y, a_1=0 )=
\probP( f(\xneg,a_1)=y, a_1=1 )\,.\nonumber
\end{equation}
\end{definition}

\begin{definition}[Disparate impact (\ie{} ``80\% rule'') \citep{feldman2015certifying,zafar2017fairness2}]
\label{def:grp1,disi}%
\begin{equation}
\topequation
    \frac{ \probP( f(\xneg,a_1)=1 \mid a_1=0 )
    }{ \probP( f(\xneg,a_1)=1 \mid a_1=1 )
    } \leqslant \tau=0.8 
    \,.\label{eq:grp,di}
\end{equation}

Its reciprocal is called the likelihood ratio positive \citep{feldman2015certifying}, given by
\begin{equation}
\topequation
\mathrm{LR}_+( f,a_1 )= \frac{
    \probP( f(\xneg,a_1)=1 \mid a_1=1 )
}{ \probP( f(\xneg,a_1)=1 \mid a_1=0 ) }
\,,\nonumber
\end{equation}
and a data set has disparate impact if 
\begin{equation}
\topequation
\mathrm{LR}_+( f,a_1 ) >\frac{1}{\tau} =1.25 
\,,\nonumber
\end{equation}
and will be convenient to work with the reciprocal of $\mathrm{LR}_+$, which is denoted by
\begin{equation}
\topequation
\mathrm{DI}= \frac{1}{ \mathrm{LR}_+( f,a_1 ) } 
\,.\nonumber
\end{equation}
\end{definition}

\begin{definition}[Disparate treatment \citep{zafar2017fairness2}, also indicated as ``statistical parity'' in \citep{corbett2017algorithmic,haas2019price}]
\label{def:grp1,dist}%
\begin{equation}
\topequation
    \probP( f(\xneg,a_1)\!=\!1 \mid a_1=j )
    = \probP( f(\xneg,a_1)\!=\!1 )
    \,,\; \forall\, j\in\{0,1\}
    \,.\label{eq:grp,dt}
\end{equation}
It is possible to apply to one multi-valued \senatt{} \citep{corbett2017algorithmic,agarwal2019fair}, called statistical parity, that is, 
\begin{equation}
\topequation
    \probP( f(\xneg,a_1)=1 \mid a_1=j )
    = \probP( f(\xneg,a_1)=1 ) 
    \,,\; \forall\, 
    j\in\mathcal{A}_1=\{1,2,...,n_{a_1}\}
    \,.\label{eq:sub,sp}
\end{equation} 
\end{definition}

\begin{definition}[Conditional statistical parity \citep{corbett2017algorithmic}]
\label{def:grp1,csp}
\begin{equation}
\topequation
\begin{split}
    \probP( f(\xneg,a_1)=1 \mid \ell(\xneg,a_1) ,\, a_1=j )
    = \probP( f(\xneg,a_1)=1 \mid \ell(\xneg,a_1) ) 
    ,\, \forall\, j\in\mathcal{A}_1=\{1,2,...,n_{a_1}\}
    \,,\label{eq:sub,cs}\nonumber
\end{split}
\end{equation}
where $\ell: \mathbb{R}^{n_d+1} \mapsto \mathbb{R}^{m}$ is a projection of features $(\xneg,a_1)$ to the factors that are considered ``legitimate''. 
\end{definition}

\begin{definition}[Bounded group loss \citep{agarwal2019fair}]
\label{def:bounded_grplos}
A predictor $h$ satisfies bounded group loss (BGL) at level $\xi$ under a distribution over $(\xneg,a_1,y)$ if 
\begin{equation}
\topequation
\probE[ \ell(y,h(\xneg,a_1)) \mid a_1=j ] 
\leqslant\xi \,,\;\forall j\in
\mathcal{A}_1=\{1,2,...,n_{a_1}\} \,,
\end{equation}
where $h(\xneg,a_1)\in[0,1]$ and 
the loss $\ell:\mathcal{Y}\times\![0,1] \mapsto[0,1]$, measuring the accuracy, is required to be $1$-Lipschitz under the $\ell_1$ norm.\footnote{
That is, $|\ell(y,u)-\ell(z,v)| \leqslant |y-z|+|u-v| \,,\; \forall y,z,u,v \,.$
} 
\end{definition}

\begin{definition}[Strategic minimax fairness \citep{diana2024minimax}]
A classifier $h\in\mathcal{H}$ satisfies ``$\gamma$-minimax fairness'' with respect to the distribution $\mathsf{D}$ if it minimises the maximum group error rate up to an additive factor of $\gamma$. In other words,
\begin{equation}
\topequation
\max_{g\in\mathcal{A}_1} \ell_g(h) \leqslant
\min_{h'\in\mathcal{H}} \max_{g\in\mathcal{A}_1} \ell_g(h')+\gamma \,.
\end{equation}
Note that the distribution $\mathsf{D}$ exists over the data domain $\mathcal{X}\times\mathcal{A}_1\times\mathcal{Y}$, and $\mathsf{D}_g$ denotes the conditional distribution of $(\mathbf{x},y)$ conditioned on the group $g\in\mathcal{A}_1$. 
Also note that the overall error rate of $h$ and its corresponding group error rate for the group $g$ are defined as follows,
\begin{small}
\begin{subequations}
\topequation
\begin{align}
\ell(h) &\defineq \mathbb{P}_{(\mathbf{x},g,y)\sim \mathsf{D}} [h(\mathbf{x},g)\neq y] \,,\\
\ell_g(h) &\defineq \mathbb{P}_{(\mathbf{x},y)\sim \mathsf{D}_g}[ h(\mathbf{x},g)\neq y ] \,.
\end{align}%
\end{subequations}%
\end{small}%
\end{definition}

\begin{definition}[Equalised odds \citep{hardt2016equality,haas2019price}]
\label{def:grp2,eodd}
\begin{equation}
\topequation
\begin{split}
    \probP( f(\xneg,a_1)=1 \mid a_1=0 ,y ) =
    \probP( f(\xneg,a_1)=1 \mid a_1=1 ,y )
    ,\, \forall\, y\in\{0,1\}
    \,.\label{eq:grp,eo'}
\end{split}
\end{equation}
\end{definition}

\begin{definition}[Equality of opportunity \citep{hardt2016equality,gajane2017formalizing,haas2019price}, \aka{} treatment equality \citep{berk2021fairness}]
\label{def:grp2,eo}
\begin{equation}
\topequation
    \probP( f(\xneg,a_1)=1 \mid a_1=0,\, y=1 )=
    \probP( f(\xneg,a_1)=1 \mid a_1=1,\, y=1 )
    \,.\label{eq:grp,eo}
\end{equation}
It has a similar form using a scoring function \citep{chouldechova2017fair}, called error rate balance, that is,
\begin{subequations}
\topequation
\begin{align}
    \probP( h(\xneg,a_1)>\thres \mid y=0,a_1=0 )=
    \probP( h(\xneg,a_1)>\thres \mid y=0,a_1=1 )\,,\nonumber\\
    \probP( h(\xneg,a_1)\leqslant\thres \mid y=1,a_1=0 )=
    \probP( h(\xneg,a_1)\leqslant\thres \mid y=1,a_1=1 )\,.\nonumber
\end{align}
\end{subequations}
It also has two other forms, that is, false positive error rate balance \citep{verma2018fairness} 
\begin{equation}
\topequation
\probP( f(\xneg,a_1)=1 \mid y=0, a_1=0 )=
\probP( f(\xneg,a_1)=1 \mid y=0, a_1=1 )\,,\nonumber
\end{equation}
and false negative error rate balance \citep{verma2018fairness}
\begin{equation}
\topequation
\probP( f(\xneg,a_1)=0\mid y=1, a_1=0 )=
\probP( f(\xneg,a_1)=0\mid y=1, a_1=1 )\,.\nonumber
\end{equation}
\end{definition}

\begin{definition}[Predictive equality \citep{corbett2017algorithmic}]
\label{def:grp2,peq}
\begin{equation}
\topequation
\begin{split}
    \probP( f(\xneg,a_1)=1 \mid a_1=j,\, y=0 ) 
    =\probP( f(\xneg,a_1)=1 \mid y=0 ) 
    ,\, \forall\, j\in\mathcal{A}_1=\{1,2,...,n_{a_1}\}
    \,.\label{eq:sub,pe}\nonumber
\end{split}
\end{equation}
\end{definition}

\begin{definition}[False positive (FP) subgroup fairness, \ie{} $\gamma$-subgroup fairness \citep{kearns2018preventing,kearns2019empirical}]
\label{def:gamma_subgroup}
\begin{equation}
\topequation
    \alpha_\text{FP}(f) \cdot
    \beta_\text{FP}(f)
    \leqslant \gamma
    \,,\label{eq:sub,sf}
\end{equation}
where
\begin{subequations}
\topequation
\begin{align}
    \alpha_\text{FP}(f) &= \probP( a_1=0, y=0 ) \,,\\
    \beta_\text{FP}(f) &=| 
    \probP( f(\xneg,a_1)=1 \mid y=0 ) 
    -\probP( f(\xneg,a_1)=1 \mid a_1=0,\, y=0 )
    | \,.
\end{align}%
\end{subequations}%
\end{definition}

\begin{definition}[Predictive parity \citep{chouldechova2017fair,verma2018fairness}, \aka{} outcome test \citep{goel2016precinct,simoiu2017problem}]
\label{def:grp3,pqp}
\begin{equation}
\topequation
    \probP( y=1\mid a_1=0,\, f(\xneg,a_1)=1 )=
    \probP( y=1\mid a_1=1,\, f(\xneg,a_1)=1 )
    \,.\label{eq:grp,pp}
\end{equation}
or in the form of a scoring function \citep{chouldechova2017fair}
\begin{equation}
\topequation
\probP( y=1\mid h(\xneg,a_1)> \thres, a_1=0 )=
\probP( y=1\mid h(\xneg,a_1)> \thres, a_1=1 ) \,.
\end{equation}
It also has another form, called conditional use accuracy equality \citep{verma2018fairness}, 
\begin{subequations}
\topequation
\begin{align}
    \probP( y=1\mid f(\xneg,a_1)=1, a_1=0 )=
    \probP( y=1\mid f(\xneg,a_1)=1, a_1=1 ) \,,\nonumber\\
    \probP( y=0\mid f(\xneg,a_1)=0, a_1=0 )=
    \probP( y=0\mid f(\xneg,a_1)=0, a_1=1 )\,.\nonumber
\end{align}
\end{subequations}
\end{definition}

\subsection{Individual fairness}
\label{subsec:lit2}

\begin{definition}[Lipschitz condition \citep{dwork2012fairness}]
\label{def:indv,lips}
A mapping/predictor $h\!: \mathcal{X\times A}_1 =\mathcal{X\times}\{0,1\} \mapsto[0,1]$ satisfies the $\lambda$-Lipschitz property if for any $(\xneg,a_1), (\xneg',a_1')$, 
\begin{equation}
    \topequation
    \dist_y( h(\xneg,a_1), h(\xneg',a_1') ) 
    \leqslant \lambda\cdot 
    \dist_x( (\xneg,a_1), (\xneg',a_1') )
    \,,\label{eq:ind,lip}
\end{equation}
where $\dist_y$ and $\dist_x$ are (task-specific) distance metrics. 
Note that $\lambda$ is a positive constant.

It can also be written as the probability Lipschitzness, \ie{}
\begin{equation}
    \topequation
    \probP\left(\frac{
        \dist_y( h(\xneg,a_1), h(\xneg',a_1') )
    }{
        \dist_x( (\xneg,a_1), (\xneg',a_1') )
    } \geqslant\epsilon
    \right) \leqslant\delta
    \,;\label{eq:ind,l1}\nonumber
\end{equation}
or the $(\epsilon-\delta)$ language formulation,
\begin{equation}
    \topequation
    \dist_x( (\xneg,a_1), (\xneg',a_1') )
    \leqslant\epsilon \Rightarrow
    \dist_y( h(\xneg,a_1), h(\xneg',a_1') )
    \leqslant\delta 
    \,,\nonumber\label{eq:ind,l2}
\end{equation}
where $\epsilon\geqslant 0$ and $\delta\geqslant 0$.

Additionally, in \citet{gajane2017formalizing}, 
a predictor satisfies individual fairness if and only if: $h(\xneg,a_1) \approx h(\xneg',a_1') \mid \dist_x((\xneg,a_1),(\xneg',a_1')) \approx 0$, where $\mathcal{X}_a\defineq \mathcal{X\!\times\! A}$ and $\dist_x: \mathcal{X}_a \times \mathcal{X}_a \mapsto\mathbb{R}$ is a distance metric for individuals.

In essence, individual fairness follows the principle that ``similar individuals should be evaluated or treated similarly.'' 
A careful choice of distance metrics is crucial in ensuring fairness \citep{luong2011k,boeschoten2021achieving}. 
\end{definition}

\begin{definition}[General entropy indices \citep{speicher2018unified} and the Theil index \citep{haas2019price}]
\label{def:indv,gei}
For a constant $\alpha\notin\{0,1\}$, the generalised entropy indices for a problem with $n$ instances are defined, to quantify algorithmic unfairness, as
\begin{equation}
    \topequation
    \mathrm{GEI}^\alpha = 
    \frac{1}{n\alpha(\alpha-1)}
    \sum_{i=1}^n \left(
        \left(\frac{b_i}{\mu}\right)^\alpha
    -1 \right) \,,
\end{equation}
where benefits $b_i= f(\xneg_i,a_{1i})-y_i+1$ and $\mu=\sfrac{\sum_i b_i}{n}$.

The Theil index is a special case for $\alpha=1$, that is, 
\begin{equation}
    \topequation
    \mathrm{Theil}= 
    \frac{1}{n}\sum_{i=1}^n
    \frac{b_i}{\mu} \log
    \left(\frac{b_i}{\mu}\right)
    \,.
\end{equation}
They are used additionally to group fairness measures to compare different algorithms and determine which one is considered the fairest from an individual perspective. 
\end{definition}

\begin{definition}[Counterfactual fairness (CFF) \citep{kusner2017counterfactual,gajane2017formalizing}]
\label{def:causal,cff}
Given a causal model\footnote{%
A causal model is defined as a triple $(\cffU,\cffV,\cffF)$ where: 
1) $\cffU$ is a set of latent background variables, unaffected by any observable variable in $\cffV$; 
2) $\cffF$ is a set of structural equations $\{f_1,...,f_m\}$, each defining a variable $\cffIV\in\cffV$ as $\cffIV=f_i(\cffPA,\mathsf{U}_{\cffPA})$, where $\cffPA\subseteq \cffV\setminus\{\cffIV\}$ (the `parents' of $\cffIV$) and $\mathsf{U}_{\cffPA}\subseteq \cffU$. 
The model is causal because, given a distribution $\mathsf{P}(\cffU)$ over the background variables $\cffU$, the distribution of any subset $\cffZ\subseteq \cffV$ can be derived after an intervention on $\cffV\setminus \cffZ$. 
An intervention on $\cffIV$ is the substitution of $\cffIV= f_i(\cffPA,\mathsf{U}_{\cffPA})$ with $\cffIV=v$ for some $v$. 
} $(\cffU,\cffV,\cffF)$ where $\cffV\equiv \mathsf{A\cup X}$, a predictor $f(\cdot)$ is \emph{counterfactual fair} if under any context $\mathsf{X}=\xneg$ and $\mathsf{A}=\xpos$,
\begin{equation}
\topequation
\begin{split}
    \probP( f_{\mathsf{A}\gets \xpos}(\mathsf{U})=y \mid \mathsf{X}=\xneg,\, \mathsf{A}=\xpos )= 
    \probP( f_{\mathsf{A}\gets \xpos'}(\mathsf{U})=y \mid \mathsf{X}=\xneg,\mathsf{A}=\xpos )
    \,,\label{eq:causal,cff}
\end{split}
\end{equation}
for all $y$ and for any value $\xpos'$ attainable by $\mathsf{A}$.

CFF is individual-level and agnostic to the accuracy of the predictor $f(\cdot)$. It substantially differs from comparing individuals who happen to share the same `treatment' $\mathsf{A}=\xpos$ and coincide on identical values of $\mathsf{X}$, as differences between $\mathsf{X}_{\mathsf{A}\gets \xpos}$ and $\mathsf{X}_{\mathsf{A}\gets \xpos'}$ must be solely caused by variations in $\mathsf{A}$. 

In addition, CFF assumes that any causal effect of $\mathsf{A}$ (a sensitive attribute) on predictions is deemed illegitimate. A broader definition \citep{nabi2018fair,boeschoten2021achieving}, where CFF is a special case, distinguishes between discriminatory and non-discriminatory causal pathways from $\mathsf{A}$. Instead of enforcing  $\mathbb{P}_{\mathrm{do}(\xpos)}(f|\mathsf{X})= \mathbb{P}_{\mathrm{do}(\xpos')}(f|\mathsf{X})$ as in CFF, fairness is achieved by inferring a distribution $\mathbb{P}^*(\mathsf{y|X})$ that closely approximates $\mathbb{P}(\mathsf{y|X})$ (in a Kullback-Leibler sense) while blocking discriminatory pathways within a given tolerance using causal inference techniques.

\end{definition}

\begin{definition}[Proxy discrimination \citep{kilbertus2017avoiding}]
\label{def:causal,pd}
Given a generic causal graph structure (a DAG), 
involving a protected attribute $\mathsf{A}_1$, a set of proxy variables $\mathsf{P}$, features $\mathsf{X}$, a predictor $f(\cdot)$, and sometimes an observed outcome $\mathsf{y}$. 
The predictor $f(\cdot)$ exhibits no \emph{individual proxy discrimination}, if for all proxies $\mathbf{p},\mathbf{p}'$,
\begin{equation}
    \topequation
    \mathbb{P}(f\mid \mathrm{do}(\mathsf{P}=\mathbf{p}) )=
    \mathbb{P}(f\mid \mathrm{do}(\mathsf{P}=\mathbf{p}') )\,,
\end{equation}
where $\mathrm{do}(\mathsf{P}=\mathbf{p})$ denotes an intervention on $\mathsf{P}$.
The proxy $\mathsf{P}$ and the features $\mathsf{X}$ could be multidimensional. 
\end{definition}

\begin{remark}
Both CFF \citep{kusner2017counterfactual,gajane2017formalizing} and proxy discrimination (PD) \citep{kilbertus2017avoiding} are grounded in causal inference, requiring an explicit causal graph to proceed with the analysis. 
This dependency limits their applicability in settings where the algorithm under investigation is not expressed as a causal model. Moreover, the definitions are not inherently quantitative, although extensions to quantitative metrics are possible, for example,
\begin{subequations}
\topequation\small
\begin{align}
    \mathrm{CFF}(f) &= 
    \max_{ \xpos,\xpos'\in\mathcal{A} ,\xpos'\neq\xpos }\{|
        \probP( f_{\mathsf{A}\gets\xpos}(\cffU)=y \mid \mathsf{X}=\xneg, \mathsf{A}=\xpos )-
        \probP( f_{\mathsf{A}\gets\xpos'}(\cffU)=y \mid \mathsf{X}=\xneg, \mathsf{A}=\xpos' )
    |\} \,,\nonumber\\
    \mathrm{PD}(f) &=
    \max_{ \mathbf{p},\mathbf{p}'\sim\mathsf{P} , \mathbf{p}'\neq\mathbf{p} }\{|
        \probP( f\mid \mathrm{do}(\mathsf{P}=\mathbf{p}), \mathsf{X}=\xneg )-
        \probP( f\mid \mathrm{do}(\mathsf{P}=\mathbf{p}'), \mathsf{X}=\xneg )
    |\} \,.\nonumber
\end{align}
\end{subequations}
Despite this apparent formalism, both CFF and PD are typically constrained to a single \senatt. When intersectional or multiple attributes are involved in CFF, they must first be preprocessed to degenerate into one ``super'' attribute before the analysis, which not only introduces a higher computational burden but also risks obscuring the nuanced ways in which intersectional identities interact with structural biases.
\end{remark}

\begin{definition}[Discriminative risk \citep{bian2023increasing_alt}]%
\label{def:my,dr}
\begin{equation}
    \topequation
    \mathrm{DR}(f)= \probE[
    \mathbb{I}( f(\xneg,\xpos)\neq f(\xneg,\xqtb) )
    ] \,,\label{eq:my,dr}
\end{equation}
where $\xqtb$ is a perturbed $\xpos$, and $n_a\geqslant 1, |\mathcal{A}_i|\geqslant 2 \,(i\in[n_a])$. 
\end{definition}

\subsection{Other distributive fairness}
\label{subsec:lit3}

\begin{definition}[Fairness through unawareness \citep{dwork2012fairness,gajane2017formalizing}]
A predictor is said to achieve \emph{fairness through unawareness (FTU)} (or unconscious/unaware fairness) if all protected attributes $\mathcal{A}$ are excluded from the decision-making process. 

Despite its compelling simplicity, this approach has a clear shortcoming: the remaining attributes $\mathcal{X}$  may contain discriminatory information analogous to $\mathcal{A}$ that may not be obvious at first, acting as \emph{proxy} attributes. As a result, discrimination cannot be guaranteed to be eliminated. 

\end{definition}

\begin{definition}[Calibration]
\label{def:calibration}
A score $h(\cdot)$ is said to be well calibrated if it reflects the same likelihood of recidivism irrespective of the individuals' group membership \citep{chouldechova2017fair}, that is to say, if for all values of $p$, 
\begin{equation}
\topequation
\probP( y=1\mid h(\xneg,a_1)=p, a_1=0 )=
\probP( y=1\mid h(\xneg,a_1)=p, a_1=1 ) \,.
\end{equation}

In \citep{pleiss2017fairness,kleinberg2016inherent}, for two binary predictors $h_1,h_0: \mathbb{R}^{n_d+1}\mapsto[0,1]$, $h_1$ classifies samples with $a_1\!=\!1$ and $h_0$ does samples with $a_1\!=\!0$. Any $h_t \,(t\in\{1,0\})$ is perfectly calibrated if 
\begin{equation}
    \topequation
    \forall p\in[0,1],\, 
    \mathbb{P}_{(\xneg,a_i=t,y)}(
        y=1\mid h_t(\xneg,a_1)=p
    )=p \,.\nonumber
\end{equation}
It intuitively prevents the probability scores from carrying group-specific information. 
\end{definition}

\begin{definition}[Multicalibration \citep{hebert2018multicalibration,gohar2023survey}]
For any individual $i\in \mathcal{X}$, some unknown probability $p_i^*\in[0,1]$ is assumed to exist, and there is no assumptions on the structure of $p^*: \mathcal{X}\mapsto[0,1]$. Particularly, enough uncertainty is assumed to exist in the outcomes that it may be hard to learn $p^*$ directly. 
Let $\mathcal{D}$ denote the distribution over individuals, supported on $\mathcal{X}$; for $D_j\subseteq \mathcal{X}$, let $i\sim D_j$ denote a sample drawn from $\mathcal{D}$ conditioned on membership in $j$ where $j\in\mathcal{A}_1$ represent a subgroup \wrt{} $a_1$. 
Then for any $\alpha>0$ and $D_j\subseteq \mathcal{X}$, 
a predictor $h$ is $\alpha$-accurate-in-expectation (AE) \wrt{} $a_1$ if 
\begin{equation}
\topequation
| \mathbb{E}_{i\in D_j}[ h_i-p_i^* ] | \leqslant \alpha 
\,.\nonumber 
\end{equation}

For any $v\in[0,1]$, $D_j\subseteq \mathcal{X}$, and predictor $h$, let $S_v= \{i: f_i=v\}$. Then for $\alpha \in [0,1]$, $h$ is $\alpha$-calibrated with respect to $D_j$ if there exists some $S'\subseteq D_j$ with $\mathbb{P}_{i\sim\mathcal{D}} [i\in S'] \geqslant (1-\alpha) \cdot \mathbb{P}_{i\sim\mathcal{D}}[i\in D_j]$ such that for all $v\in[0,1]$,
\begin{equation}
\topequation
| \mathbb{E}_{i\sim S_v \cap S'} [h_i-p_i^*] | \leqslant\alpha
\,;\nonumber
\end{equation}
Note that $\alpha$-calibration \wrt{} $D_j$ implies $2\alpha$-AE \wrt{} $D_j$.

Let $\mathcal{C}\subseteq 2^\mathcal{X}$ be a collection of subsets of $\mathcal{X}$ and $\alpha\in[0,1]$. 
A predictor $h$ is $(\mathcal{C},\alpha)$-multicalibrated if for all $S\in\mathcal{C}$, $h$ is $\alpha$-calibrated with respect to $S$. 
\end{definition}

\begin{definition}[Multiaccuracy \citep{kim2019multiaccuracy}]
Let $\alpha\geqslant 0$ and let $\mathcal{C}\subseteq [-1,1]^\mathcal{X}$ be a class of functions on $\mathcal{X}$. A hypothesis $f:\mathcal{X}\mapsto[0,1]$ is $(\mathcal{C},\alpha)$-multiaccurate if for all $c\in\mathcal{C}$:
\begin{equation}
\topequation
| \probE[ c(\xneg,a_1)\cdot( f(\xneg,a_1)-y ) ] |
\leqslant\alpha \,.
\end{equation}
$(\mathcal{C},\alpha)$-multiaccuracy guarantees that a hypothesis appears unbiassed according to a class of statistical tests defined by $\mathcal{C}$. 
By defining the class in terms of a collection of subsets $S_j\subseteq \mathcal{X}$ conditioned on membership, and taking $\mathcal{C}$ to be 
$\indicator(a_1=j\mid (\xneg,a_1))$ (and its negation) for each subset in the collection, $(\mathcal{C},\alpha)$-multiaccuracy guarantees that for each $S_j$, the predictions of $f$ are at most $\alpha$-biassed. 
\end{definition}

\begin{definition}[Differentially fair and intersectionality \citep{foulds2020intersectional}]
Considering several discrete-valued protected attributes, that is, $\mathcal{A}= \mathcal{A}_1\times \mathcal{A}_2\times ...\times \mathcal{A}_{n_a}$, 
a predictor $f(\cdot)$ is $\epsilon$-differentially fair (DF) if with respect to $(\mathcal{A},\mathcal{D})$ if for all $\theta\in\mathcal{D}$ with $(\xneg,\xpos)\sim \theta$, 
\begin{equation}
\topequation
e^{-\epsilon} \leqslant
\frac{ \probP(f(\xneg,\xpos)=y \mid \xpos=\xpos') }{ \probP(f(\xneg,\xpos)=y \mid \xpos=\xpos'') }
\leqslant e^{\epsilon} \,,\nonumber
\end{equation}
for all $(\xpos',\xpos'')\in \mathcal{A}\times\mathcal{A}$ where $\probP(\xpos'\mid \theta)>0$ and $\probP(\xpos''\mid \theta)>0$. 
And the empirical differential fairness (EDF) corresponds to verifying that for any $y,\xpos',\xpos''$, we have
\begin{equation}
\topequation
e^{-\epsilon} \leqslant
\frac{ N_{y,\xpos'} }{ N_{\xpos'} }
\frac{ N_{\xpos''} }{ N_{y,\xpos''} }
\leqslant e^{\epsilon} 
\,,\nonumber
\end{equation}
where $N_{y,\xpos}$ and $N_{\xpos}$ are empirical counts of their subscripted values in the dataset $D$. 
It has an intersectionality property: let $f$ be an $\epsilon$-differentially fair predictor in $(\mathcal{A},\mathcal{D})$, and let $\mathcal{B}= \mathcal{A}_a\times \mathcal{A}_b\times ...\times \mathcal{A}_k$ be the Cartesian product of a nonempty proper subset of the protected attributes included in $\mathcal{A}$, then $f$ is $\epsilon$-differentially fair in $(\mathcal{B},\mathcal{D})$. 
\end{definition}

\begin{definition}[Group benefit ratio and worst-case min-max ratio \citep{ghosh2021characterizing}]
Considering several discrete-valued protected attributes $\mathcal{A}= \mathcal{A}_1\times \mathcal{A}_2\times ...\times\mathcal{A}_{n_a}$, 
group benefit equality \citep{gartner2025new}, aiming to be useful in the domain of healthcare, measures the predicted rate of passing for a subgroup compared to the actual rate of passing, that is,
\begin{equation}
\topequation
\probP( f(\xneg,\xpos) \mid \xpos=\xpos' )=
\probP( y\mid \xpos=\xpos' )\,,
\;\forall\, \xpos'\in\mathcal{A}.\nonumber
\end{equation}
Essentially, it is treating $\mathcal{A}$ as a super protected attribute with $n_{a_1}\times n_{a_2}\times ...\times n_{n_a}$ possible values, that is
\begin{equation}
\topequation
\probP( f(\xneg,a_1)=1 \mid a_1=j )=
\probP( y=1 \mid a_1=j ) \,,
\;\forall\, j\in\mathcal{A}_1 \,.
\end{equation}

Group benefit ratio for one subgroup is defined as
\begin{equation}
\topequation
\mathrm{GBR}_j =\frac{
    \probP( f(\xneg,a_1)=1 \mid a_1=j )
}{  \probP( y=1 \mid a_1=j ) } \,,\nonumber
\;\forall\, j\in\mathcal{A}_1 \,,
\end{equation}
and using the worst case, min-max ratio definition, group benefit ratio ($\mathrm{GBR\_INT}$) is defined intersectionally as
\begin{equation}
\topequation
\mathrm{GBR\_INT} =\frac{
    \min\{ \mathrm{GBR}_j ,
    \forall j\in\mathcal{A}_1 \}
}{
    \max\{ \mathrm{GBR}_j ,
    \forall j\in\mathcal{A}_1 \}
} \,.
\end{equation}

\citet{ghosh2021characterizing} also discussed other group fairness metrics and their worst case min-max ratio definition, that is, demographic parity ratio (DPR), conditional statiscial parity ratio (CSPR), and equal opportunity ratio (EOppR), given by
\begin{subequations}
\topequation
\begin{align}
\mathrm{DPR} &= \frac{
    \min\{ \probP(f(\xneg,a_1)=1 \mid a_1=j) 
    ,\,\forall j\in\mathcal{A}_1 \} 
}{
    \max\{ \probP(f(\xneg,a_1)=1 \mid a_1=j) 
    ,\,\forall j\in\mathcal{A}_1 \} 
} \,,\nonumber\\
\mathrm{CSPR} &= \frac{
    \min\{ \probP( f(\xneg,a_1)=1\mid \ell(\xneg,a_1)=1,a_1=j ) 
    ,\, \forall j\in\mathcal{A}_1 \}
}{
    \max\{ \probP( f(\xneg,a_1)=1\mid \ell(\xneg,a_1)=1,a_1=j )
    ,\, \forall j\in\mathcal{A}_1 \}
} \,,\nonumber\\
\mathrm{EOppR} &= \frac{
    \min\{ \probP( f(\xneg,a_1)=1 \mid a_1=j,y=1 ) 
    ,\, \forall j\in\mathcal{A}_1 \}
}{
    \max\{ \probP( f(\xneg,a_1)=1 \mid a_1=j,y=1 ) 
    ,\, \forall j\in\mathcal{A}_1 \}
} \,,\nonumber
\end{align}
\end{subequations}
where $\ell$ is a set of legitimate attributes,
as well as intersectional disparate impact (IDI), that is,
\begin{equation}
\topequation
\mathrm{IDI} = \min\left\{ \frac{
    \probP( f(\xneg,a_1)=1 \mid a_1=j_a ) 
}{
    \probP( f(\xneg,a_1)=1 \mid a_1=j_b ) 
} \,,\;\forall\, j_a,j_b\in 
\mathcal{A}_1, j_a\neq j_b
\right\} \,.\nonumber
\end{equation}
\end{definition}

\begin{definition}[Intersectional fairness without demographics \citep{gohar2023survey}]
The fairness objective to minimise the worst-case loss can be formulated as
\begin{equation}
\topequation
\mathcal{L}_{\max}(f) =
\max_{a_i\in \mathcal{A}}
\probE[ \ell(f;X) ] \,,\nonumber
\end{equation}
where $\probE[\ell(f;X)]$ is the expected loss for a loss function \eg{} log loss. 
Note that it is not used to evaluate unfairness level of ML models, in other words, not a fairness measure. 
\end{definition}

\begin{definition}[Harmonic fairness via manifold \citep{bian2024does,bian2024approximating}]
\label{def:my,hfm} 
Given a dataset $D=(\mathsf{X,A,Y})$, it has three versions: (1) the \emph{previous} HFM for one bi-valued \senatt, and (2) the \emph{maximal (resp. average)} HFM for several multi-valued \sapl. 

For one bi-valued SA $a_1\in \mathcal{A}_1 =\{0,1\}$, $D$ is divided into $D_1= \{ (\xneg_a,y)\defineq (\xneg,a_1,y)\in D\mid a_1=1\}$ and $\bar{D}_1= D\setminus D_1$, then given a specific distance metric $\dist(\cdot,\cdot)$ (\eg{} the standard Euclidean metric), the previous HFM is 
\begin{equation}
\topequation
    \mathrm{HFM}_\text{prev}(f) =
    \frac{ \disf_f(D_1,\bar{D}_1) }{ \disf(D_1,\bar{D}_1) }
    -1 \,,
\end{equation}
where 
\begin{equation}
\topequation%
\begin{split}
    \disf_\cdot( D_1,\bar{D}_1;\newy )=\max\{
    & \textstyle 
    \max_{(\xneg_a,y)\in D_1} \min_{(\xneg_a',y')\in \bar{D}_1} \dist((\xneg,\newy),(\xneg',\newy')) 
    ,\, \\
    & \textstyle
    \max_{(\xneg_a',y')\in \bar{D}_1} \min_{(\xneg_a,y)\in D_1} \dist((\xneg,\newy),(\xneg',\newy'))
    \} ,\nonumber 
\end{split}
\end{equation}
and $\disf_f(D_1,\bar{D}_1)= \disf_\cdot( D_1,\bar{D}_1;f(\xneg,a_1) )$ and $\disf(D_1,\bar{D}_1)\!=\! \disf_\cdot( D_1,\bar{D}_1;y )$ are two abbreviations for brevity.

For one or more multi-valued \sapl{} $\xpos\in\mathcal{A}$ where $n_a\geqslant 1$ and $|\mathcal{A}_i|\geqslant 2 \,(i\in[n_a])$, 
the maximal (resp. average) HFM are 
\begin{subequations}
\topequation
\begin{align}
    \mathrm{HFM}(f) &= \log\left(
    \frac{ \disf_{f,\xpos}(D) }{ \disf_{\xpos}(D) }
    \right) \,,\\
    \mathrm{HFM}^\text{avg}(f) &= \log\left(
    \frac{ \disf_{f,\xpos}^\text{avg}(D) }{ \disf_{\xpos}^\text{avg}(D) }
    \right) \,,
\end{align}%
\end{subequations}%
where
\begin{subequations}
\topequation%
\begin{align}
    \disf_{\cdot,\xpos}(D;\newy) &= 
    \max_{1\leqslant i\leqslant n_a} 
    \disf_{\cdot,\xpos}(D,a_i;\newy) \,,\\
    \disf_{\cdot,\xpos}^\text{avg}(D;\newy) &=
    \textstyle
    \frac{1}{n_a}\sum_{i=1}^{n_a}
    \disf_{\cdot,\xpos}^\text{avg}(D,a_i;\newy) \,,\\
    \disf_{\cdot,\xpos}(D,a_i;\newy) &=\textstyle
    \max_{j\in \{1,2,...,n_{a_i}\} }\{
        \max_{(\xneg_a,y)\in D_j}
        \min_{(\xneg_a',y')\in \bar{D}_j} 
        \dist((\xneg,\newy),(\xneg',\newy'))
    \},\nonumber\\
    \disf_{\cdot,\xpos}^\text{avg}(D,a_i;\newy) &=
    \textstyle \frac{1}{n} 
    \sum_{j\in \{1,2,...,n_{a_i}\} }
    \sum_{(\xneg,y)\in D_j}
    \min_{(\xneg',y')\in \bar{D}_j}
    \dist((\xneg,\newy),(\xneg',\newy'))
    .\nonumber
\end{align}%
\end{subequations}%
Note that $D_j= \{(\xneg_a,y)\in D\mid a_i=j \} ,\, \bar{D}_j= D\setminus D_j,$ and special case $\disf_{\cdot,\xpos}(D,a_i;\newy) = \disf_\cdot(D_1,\bar{D}_1;\newy)$ when $\mathcal{A}_i = \{0,1\}$. 
\end{definition}

\begin{definition}[$\epsilon$-BER (balanced error rate) fairness \citep{feldman2015certifying}]
Let $f': \mathcal{X\mapsto A}_1$ be a predictor of $a_1$ from $\xneg$, then the \emph{balanced error rate} (BER) of $f'$ on distribution $\mathcal{D}$ over the pair $(\xneg,a_1)$ is defined as the (unweighted) average class-conditioned error of $f'$, that is,
\begin{equation}
\topequation%
\begin{split}
    \mathrm{BER}( f'(\xneg),a_1 )\defineq
    \tfrac{1}{2} [  
        \probP( f'(\xneg)=0 \mid a_1=1 )
        + \probP( f'(\xneg)=1 \mid a_1=0 )
    ]\nonumber \,.
\end{split}
\end{equation} 
Then $a_1$ is said to be \emph{$\epsilon$-predictable} from $\xneg$ if there exists a function $f':\mathcal{X\mapsto A}_1$ such that
\begin{equation}
\topequation
\mathrm{BER}( f(\xneg),a_1 ) \leqslant\epsilon
\,,\nonumber
\end{equation}
and a data set $D=(X,A_1,Y)$ is said to be \emph{$\epsilon$-fair} if for any classification algorithm $f':\mathcal{X\mapsto A}_1$, 
\begin{equation}
\topequation
    \mathrm{BER}(f'(\xneg),a_1) 
    >\epsilon \,,\nonumber
\end{equation}
with (empirical) probabilities estimated from $D$. 
Note that a data set is \emph{not} predictable. 
Also note that, unlike other fairness definitions or measures, it is \emph{not} used to assess predictors. 

\end{definition}

\subsection{Procedural fairness \citep{grgic2016case,grgic2018beyond}}
\label{subsec:lit4}
Here we use $D$ to denote the set of all instances/members (or queried users of society), and $S$ the set of all possible features that might be used in the decision-making process (in other words, $|S| \leqslant n_d+n_a$). 
Given a set of features $S'$, let $f_{S'}$ denote the classifier that uses those features $S'$. 

\begin{definition}[Feature-apriori fairness]
\label{def:procedural,1}
For a given feature $s\in S$, let $D_s\subseteq D$ denote the set of all members that consider the feature $s$ fair to use \emph{without a priori knowledge} of how its usage affects outcomes. 
Then
\begin{equation}
    \topequation
    \mathrm{PF}_\text{apr}(f_{S'}) \defineq
    \frac{| \bigcap_{s_i\in S'} D_{s_i} |}{|D|}
    \,.\label{eq:pcd,1}
\end{equation}
\end{definition}

\begin{definition}[Feature-accuracy fairness]
\label{def:procedural,2}
Let $D_s^\text{acc} \subseteq D$ denote the set of all members that consider the feature $s$ fair to be use \emph{if it increases the accuracy of the classifier}. 
Note that typically $D_s \!\subseteq\! D_s^\text{acc}$ is expected, though this need not always hold exactly (due to either noise in estimating member preferences, or some members attaching some sort of negative connotation to the notion of accuracy). 
Then
\begin{equation}
    \topequation%
    \mathrm{PF}_\text{acc}(f_{S'}) \defineq
    \frac{| \bigcap_{s_i\in S'} \mathrm{Cond}(D_{s_i}, D_{s_i}^{\text{acc}}) |}{|D|}
    \,,\label{eq:pcd,2}
\end{equation}
where
\begin{equation*}
    \topequation
    \mathrm{Cond}( D_{s_i}, D_{s_i}^{\text{acc}} )
    = \begin{cases}
        D_{s_i} ,\hspace{1.4em}&\text{ if } \acc(f_{S'})\leqslant \acc(f_{S'\setminus\{s_i\}}) \,;\\
        D_{s_i} \bigcup D_{s_i}^\text{acc} =D_{s_i}^\text{acc} ,&\text{ otherwise} \,.\\
    \end{cases}
\end{equation*}
\end{definition}

\begin{definition}[Feature-disparity fairness]
\label{def:procedural,3}
Let $D_s^\text{disp} \subseteq D$ denote the set of all members that consider the feature $s$ fair to use \emph{even if it increases a measure of disparity} (\ie{} disparate impact or disparate mistreatment) of the classifier. 
Typically $D_s^\text{disp} \subseteq D_s$ is expected, though this need not always hold strictly due to estimation error or other reasons. 
Let $\mathrm{disp}(f_{S'})$ denote the disparity it induces, and then
\begin{equation}
    \topequation
    \mathrm{PF}_\text{disp}(f_{S'}) \defineq
    \frac{| \bigcap_{s_i\in S'} \mathrm{Cond}( D_{s_i},D_{s_i}^\text{disp} ) |}{|D|}
    \,,\label{eq:pcd,3}
\end{equation}
where
\begin{equation*}
    \topequation
    \mathrm{Cond}( D_{s_i},D_{s_i}^\text{disp} )
    = \begin{cases}
        D_{s_i}^\text{disp} ,\hspace{.9em}&\text{ if } \mathrm{disp}(f_{S'})> \mathrm{disp}(f_{S'\setminus\{s_i\}}) \,;\\
        D_{s_i} \bigcup D_{s_i}^\text{disp} =D_{s_i} ,&\text{ otherwise} \,.
    \end{cases}
\end{equation*}
\end{definition}

These three measures accommodate scenarios with multiple \sapl, each potentially having multiple values. Despite this advantage, they rely heavily on features and on a set of members/users who perceive these features as fair, which may still introduce hidden discrimination or human prejudice. Moreover, their computation is complex and time-consuming, as user judgements may evolve with learning, limiting their practical applicability. 
Additionally, \citet{wang2024procedural} proposed an FAE-based (feature attribution explanation) metric to assess group procedural fairness, which depends on the specific FAE techniques employed.

\begin{definition}[FAE-based group procedural fairness \citep{wang2024procedural}]
\label{def:procedural,4}
A given dataset $D$ is divided into two subsets by the values of a single \senatt, that is, $D_1=\{ (\mathbf{x}_i,a_{1i},y_i)\in D| a_{1i}=1 \}$ and $D_2=\{ (\mathbf{x}_i,a_{1i},y_i)\in D| a_{1i}=0 \}$. 
A local FAE function $g(\cdot)$ takes a model $f(\cdot)$ and an explained data point $(\mathbf{x}_i,a_{1i})$ as inputs and returns explanations (\ie{} feature importance scores) $\mathbf{e}_i= g(f,\mathbf{x}_i,a_{1i}) \in\mathbb{R}^{n_d+1}$, where its $j$-th component $e_{ij}$ is the importance score of the feature $x_{ij}$ for the model's prediction $f(\mathbf{x}_i,a_{1i})$. 
For a distance measure $d_e(\cdot,\cdot)$ between two sets of FAE explanation results $E_1$ and $E_2$, then
\begin{equation}
\topequation
\begin{split}
\mathrm{GPF}_\text{FAE} &= d_e(E_1,E_2) \,;\\
E_1 &= \{ \mathbf{e}_i \mid
    \mathbf{e}_i= g(f,\mathbf{x}_i,a_{1i}), 
    (\mathbf{x}_i,a_{1i})\in D_1^\prime
\}\,,\\
E_2 &= \{ \mathbf{e}_j \mid
    \mathbf{e}_j= g(f,\mathbf{x}_j,a_{1j}), 
    (\mathbf{x}_j,a_{1j})\in D_2^\prime
\}\,,
\end{split}%
\end{equation}%
where $D_1^\prime$ and $D_2^\prime$ are sets of $n$ data points from $D_1$ and $D_2$, respectively, generated by \citep[Algorithm 1]{wang2024procedural}. 
\end{definition}

\section{Experimental setup}
\label{sec:expert}

In this section, we elaborate on our experimental settings to evaluate existing fairness measures and their possible extensions. 

\paragraph{Datasets} 
We collected five public datasets but mainly used three of them in the experiments, because Ricci and Credit only have bi-valued sensitive attributes. 
Detailed information about them are provided in Table~\ref{tab:data}.

\begin{table}[h]
\caption{Data statistics. 
}\label{tab:data}
\renewcommand\tabcolsep{1.74pt}%
\begin{tabular*}{\textwidth}{@{\extracolsep\fill}rrrrrrrrrr}%
\toprule%
\multicolumn{1}{l}{\bf Dataset} & \textbf{\#inst}\footnotemark[1] & \multicolumn{2}{@{}c@{}}{\bf \#feat\footnotemark[1]} 
& \multicolumn{3}{@{}c@{}}{\bf 1st sen-att\footnotemark[2]} & \multicolumn{3}{@{}c@{}}{\bf 2nd sen-att\footnotemark[2]} \\
\cmidrule{3-4}\cmidrule{5-7}\cmidrule{8-10}%
& & raw & prep & & \#val & \#in-priv & & \#val & \#in-priv \\
\midrule
Ricci  \citep{dataset1_2024} &   118 &  5 &   6 & 
race& 3 & 68 & --- & --- & --- \\
Credit \citep{dataset2_2024} &  1000 & 20 &  58 & 
sex & 2 & 690 & age & 2 & 851 \\
Income \citep{dataset3_2024} & 30162 & 13 &  98 & 
race& 5 & 25933 & sex & 2 & 20380 \\
Compas PPR  \citep{dataset4_2024} &  6167 & 10 & 401 & 
sex & 2 & 4994 & race & 6 & 2100 \\
Compas PPVR \citep{dataset4_2024} &  4010 & 10 & 327 & 
sex & 2 & 3173 & race & 6 & 1452 \\
\botrule
\end{tabular*}
\footnotetext[1]{The columns `\#inst' and `\#feat' represent the number of instances and the number of features (including one or two sensitive attributes, but excluding classification labels), respectively. Note that `prep' is the number of features after preprocessing.}
\footnotetext[2]{For each sensitive attribute (sen-att), `\#val' and `\#in-priv' mean the number of its values and the number of members in the privileged group, respectively.}
\end{table}

\paragraph{Evaluation metrics} 
We consider accuracy and $\mathrm{f}_1$ score as performance metrics. 
We also consider the geometric mean \citep{akosa2017predictive} because data imbalance usually occurs within the datasets relevant to discrimination/bias mitigation. 
We directly use the time cost as the efficiency metric. 
As for fairness measures, we choose three commonly used group fairness measures, some individual fairness measures, as well as harmonic fairness via manifold \citep{bian2024does,bian2024approximating}: 
(1) the three commonly used group fairness measures are demographic parity (DP) \citep{feldman2015certifying,gajane2017formalizing}, equality of opportunity (EOpp) \citep{hardt2016equality}, and predictive parity (PP) \citep{chouldechova2017fair,verma2018fairness}; 
(2) the individual fairness measures used are general entropy indices (GEI) \citep{speicher2018unified} and the Theil index \citep{haas2019price}; and
(3) two other measures that can assess fairness from both individal and group aspects, that is, discriminative risk (DR) \citep{bian2023increasing_alt} and harmonic fairness via manifold (HFM, with three versions) \citep{bian2024does,bian2024approximating}.

\paragraph{Implementation details}
We mainly use bagging, AdaBoost, LightGBM \citep{ke2017lightgbm}, FairGBM \citep{cruz2022fairgbm}, and AdaFair \citep{iosifidis2019adafair} as learning algorithms, where FairGBM and AdaFair are two fairness-aware ensemble-based methods. 
Standard 5-fold cross-validation is used in these experiments; in other words, in each iteration, the entire dataset is divided into two parts, with 80\% as the training set and 20\% as the test set. 
Also, features of datasets are scaled in preprocessing to lie between 0 and 1.

\section{Additional empirical results}
\label{sec:appx}
We include more empirical results here to save space, that is, Figures~\ref{fig:radar,ppr} to \ref{fig:rel,contd1}.

\begin{figure}[tb]\centering
\begin{minipage}{\textwidth}\centering
\subfloat[]{\includegraphics[height=3.2cm]{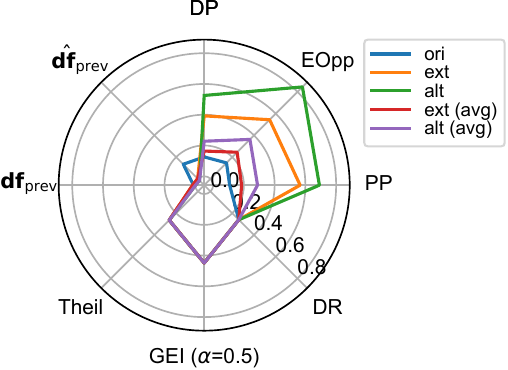}}
\subfloat[]{\includegraphics[height=3.2cm]{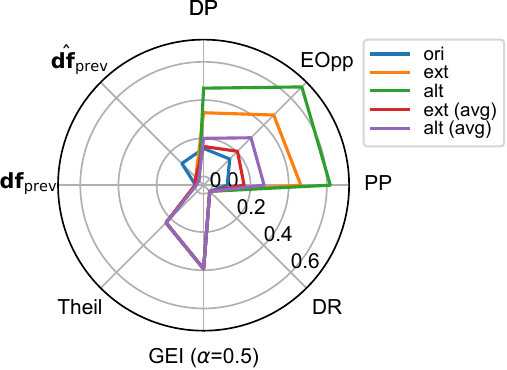}}
\subfloat[]{\includegraphics[height=3.2cm]{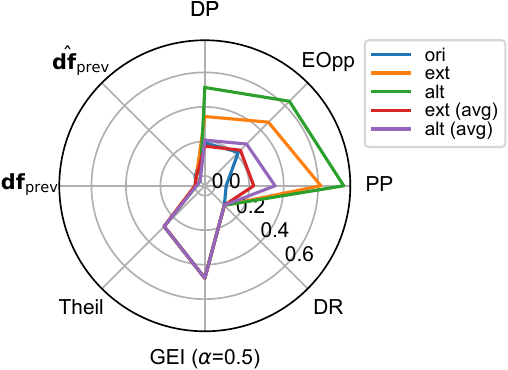}}
\vspace{-3mm}\\
\subfloat[]{\includegraphics[height=3.2cm]{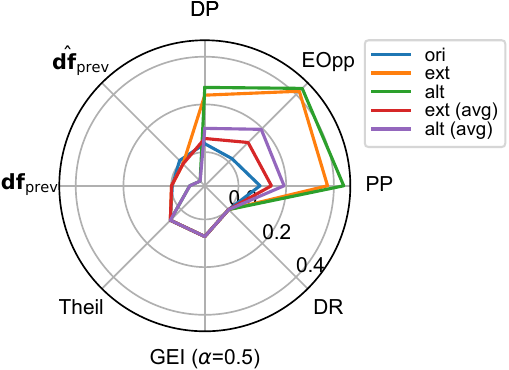}}
\subfloat[]{\includegraphics[height=3.2cm]{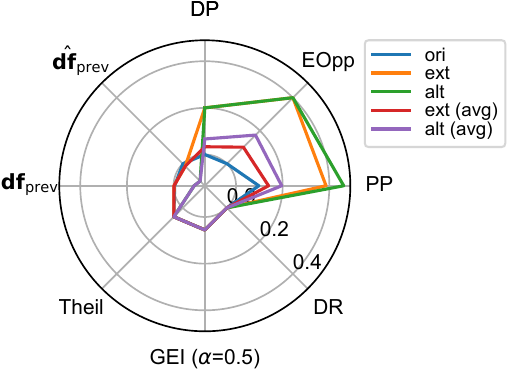}}
\subfloat[]{\includegraphics[height=3.2cm]{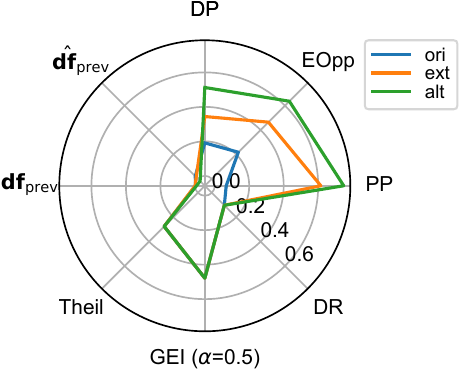}}
\hspace{1mm}
\vspace{-1mm}\caption{\small%
Comparison of fairness measures between their original definisions and their corresponding extension formulas, on the Compas PPR dataset. 
The meaning of the used notations can refer to Figure~\ref{fig:radar,adult}. 
(a--e) Using bagging, AdaBoost, LightGBM, AdaFair (trained using the first \senatt), and AdaFair (trained using the second \senatt), respectively; (f) Using LightGBM.
}\label{fig:radar,ppr}
\vspace{-4mm}
\end{minipage}
\begin{minipage}{\textwidth}\centering
\subfloat[]{\includegraphics[height=3.2cm]{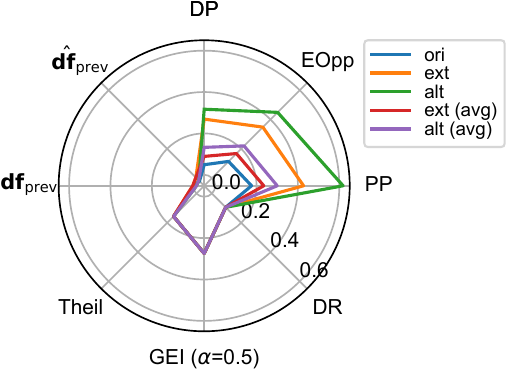}}
\subfloat[]{\includegraphics[height=3.2cm]{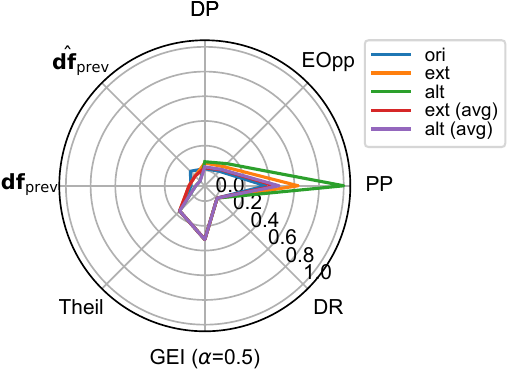}}
\subfloat[]{\includegraphics[height=3.2cm]{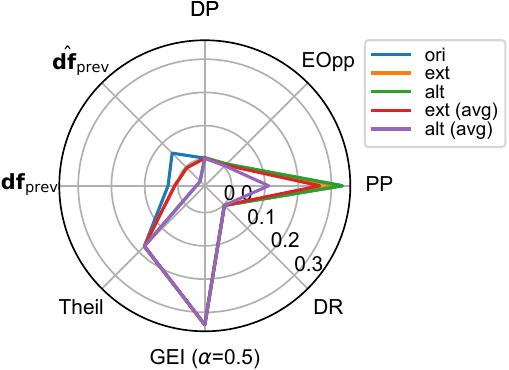}}
\vspace{-3mm}\\
\subfloat[]{\includegraphics[height=3.2cm]{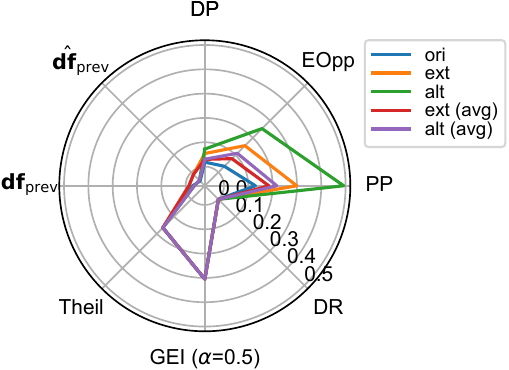}}
\subfloat[]{\includegraphics[height=3.2cm]{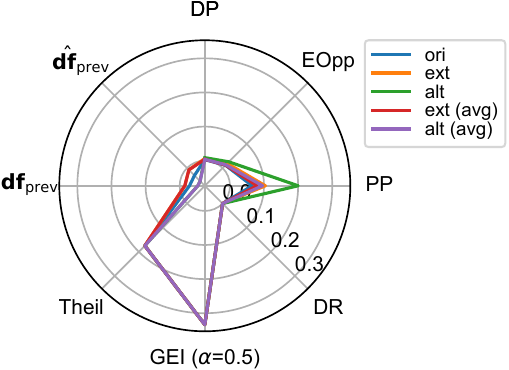}}
\subfloat[]{\includegraphics[height=3.2cm]{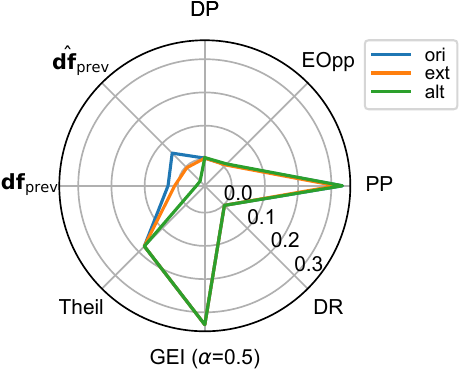}}
\hspace{1mm}
\vspace{-1mm}\caption{\small%
Comparison of fairness measures between their original definisions and their corresponding extension formulas, on the Compas PPVR dataset. 
The meaning of the used notations can refer to Figure~\ref{fig:radar,adult}. 
(a--e) Using bagging, AdaBoost, LightGBM, AdaFair (trained using the first \senatt), and AdaFair (trained using the second \senatt), respectively; (f) Using LightGBM.
}\label{fig:radar,ppvr}
\end{minipage}
\end{figure}

\begin{figure}[tb]
\centering%
\subfloat[]{\label{subfig,toc,a}%
\includegraphics[height=3.3cm]{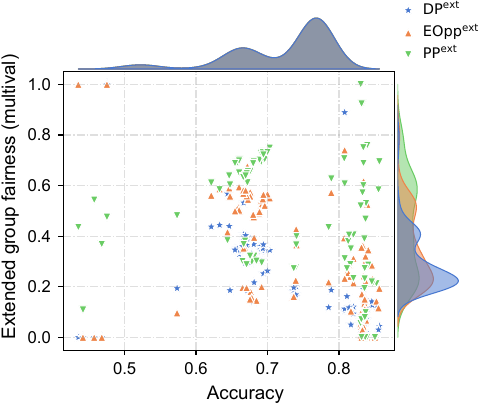}}\hspace{3mm}
\subfloat[]{\label{subfig,toc,b}%
\includegraphics[height=3.3cm]{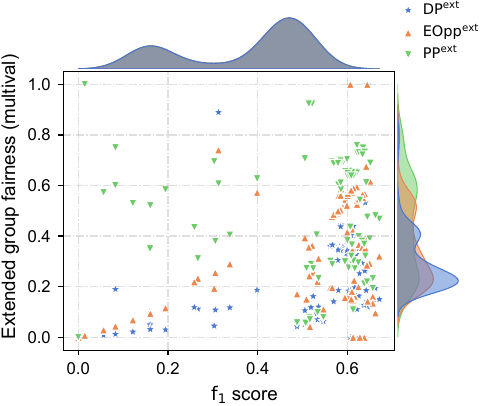}}\hspace{3mm}
\subfloat[]{\label{subfig,toc,c}%
\includegraphics[height=3.3cm]{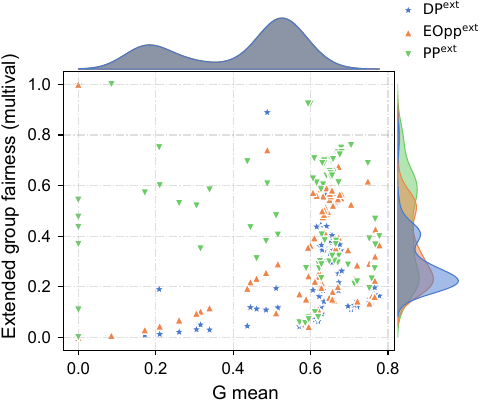}}
\vspace{-3mm}\\
\subfloat[]{\label{subfig,toc,d}%
\includegraphics[height=3.3cm]{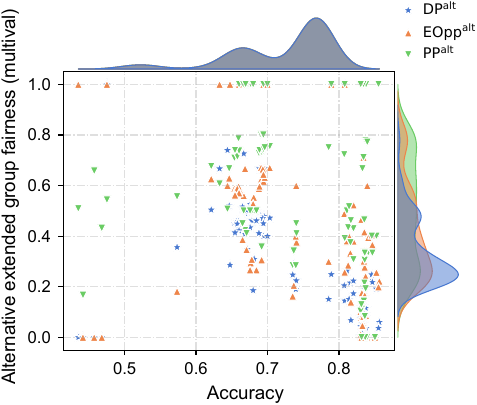}}\hspace{3mm}
\subfloat[]{\label{subfig,toc,e}%
\includegraphics[height=3.3cm]{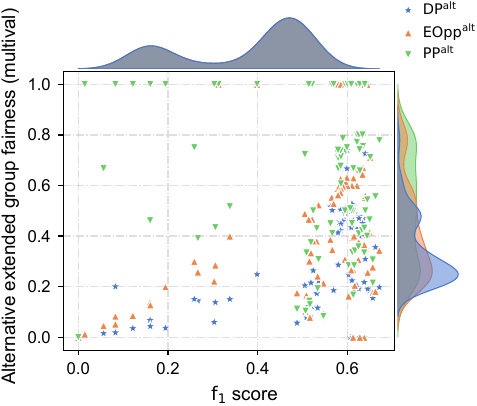}}\hspace{3mm}
\subfloat[]{\label{subfig,toc,f}%
\includegraphics[height=3.3cm]{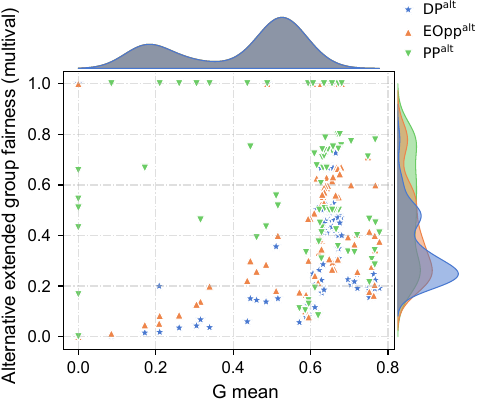}}
\vspace{-3mm}\\
\subfloat[]{\label{subfig,toc,g}%
\includegraphics[height=3.3cm]{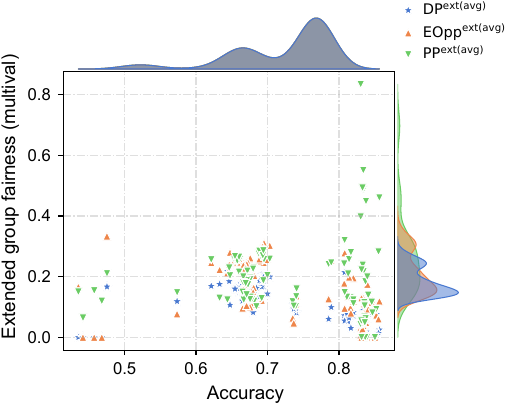}}\hspace{1mm}
\subfloat[]{\label{subfig,toc,h}%
\includegraphics[height=3.3cm]{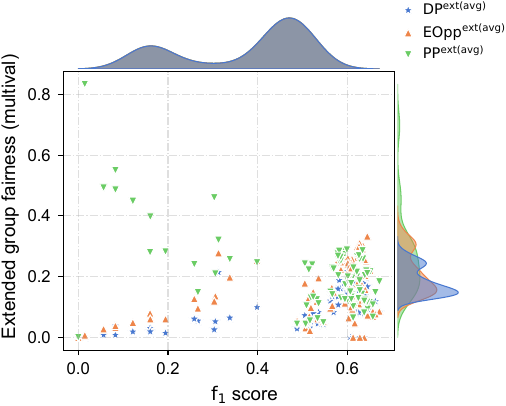}}\hspace{1mm}
\subfloat[]{\label{subfig,toc,i}%
\includegraphics[height=3.3cm]{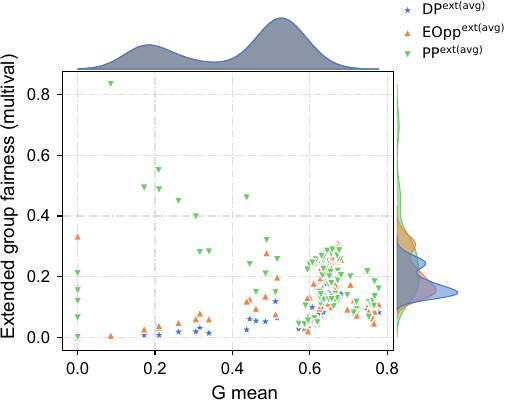}}
\vspace{-3mm}\\
\subfloat[]{\label{subfig,toc,j}%
\includegraphics[height=3.3cm]{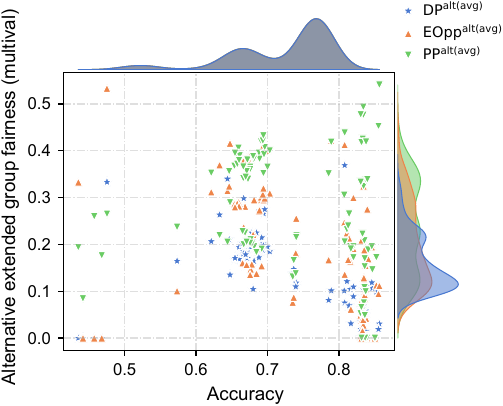}}\hspace{1mm}
\subfloat[]{\label{subfig,toc,k}%
\includegraphics[height=3.3cm]{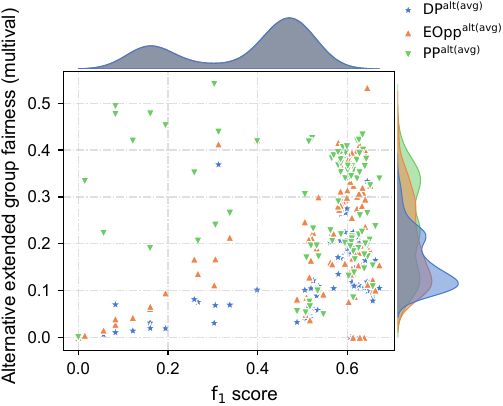}}\hspace{1mm}
\subfloat[]{\label{subfig,toc,l}%
\includegraphics[height=3.3cm]{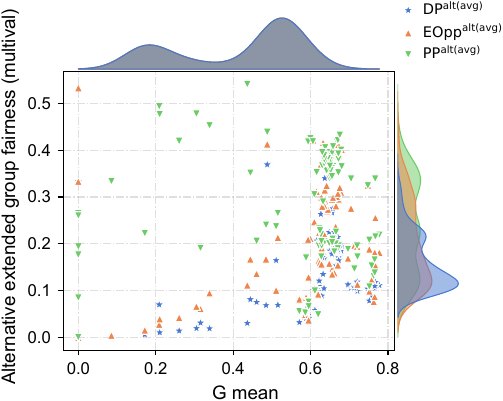}}
\vspace{-1mm}\caption{\small
Continuation of Figure~\ref{fig:trade-off}, that is, 
scatter plot between performance (accuracy, $\mathrm{f}_1$ score, or geometric mean \citep{akosa2017predictive}, respectively) and fairness. 
(a--c) Using \eqref{eqn,grp1,ext} and its analogous formulas; 
(d--f) Using \eqref{eqn,grp1,alt} and its analogous formulas; 
(g--i) Using \eqref{eqn,grp1,ext,avg} and its analogous formulas; 
(j--l) Using \eqref{eqn,grp1,alt,avg} and its analogous formulas. 
}\label{fig:t/o,cont}%
\end{figure}

\begin{figure}[tb]\centering
\subfloat[]{\label{subfig:rel,a}%
\includegraphics[height=3.4cm]{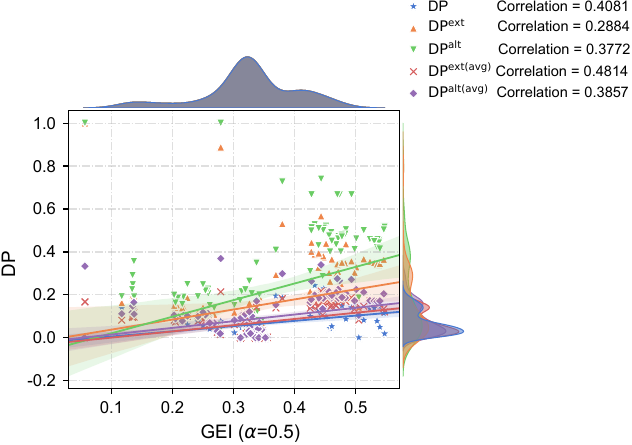}}
\subfloat[]{\label{subfig:rel,b}%
\includegraphics[height=3.4cm]{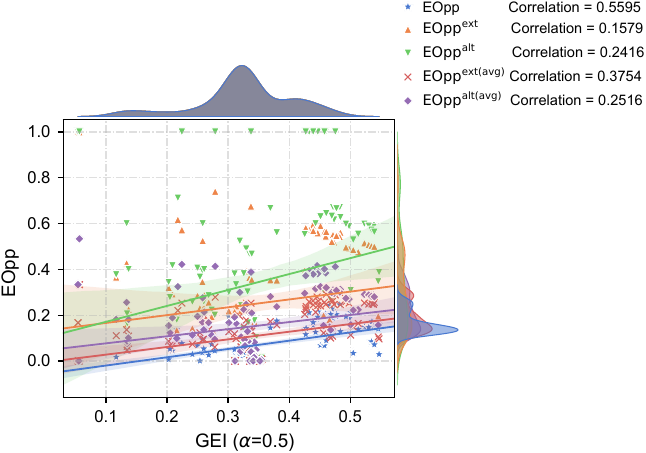}}
\subfloat[]{\label{subfig:rel,c}%
\includegraphics[height=3.4cm]{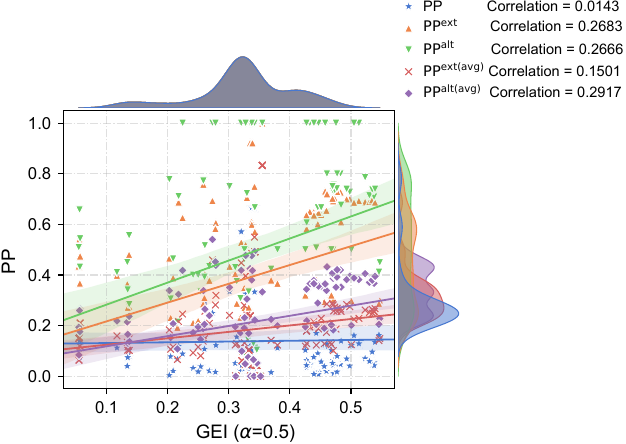}}
\vspace{-3mm}\\
\subfloat[]{\label{subfig:rel,cd1,g}%
\includegraphics[height=3.4cm]{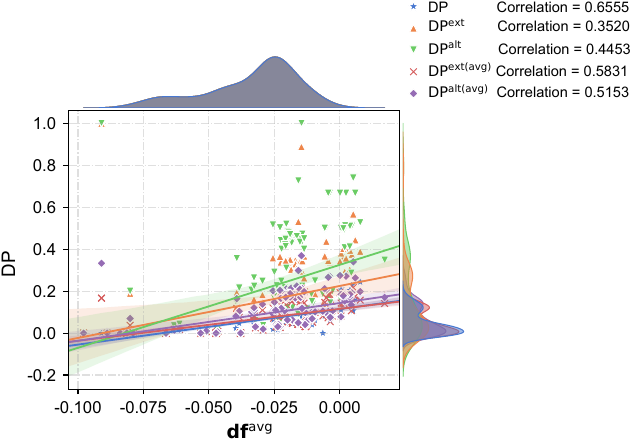}}
\subfloat[]{\label{subfig:rel,cd1,h}%
\includegraphics[height=3.4cm]{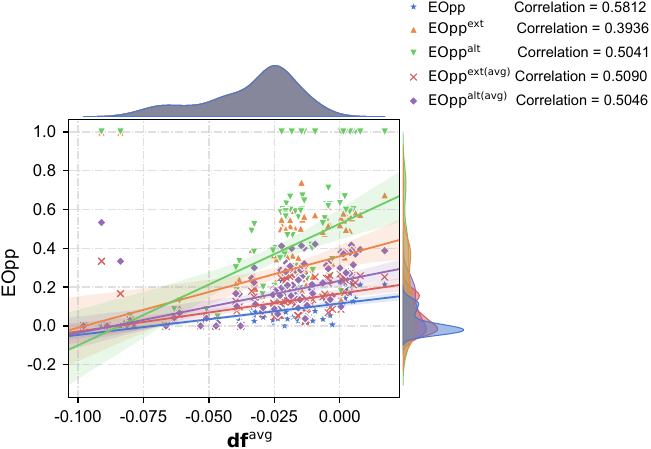}}
\subfloat[]{\label{subfig:rel,cd1,i}%
\includegraphics[height=3.4cm]{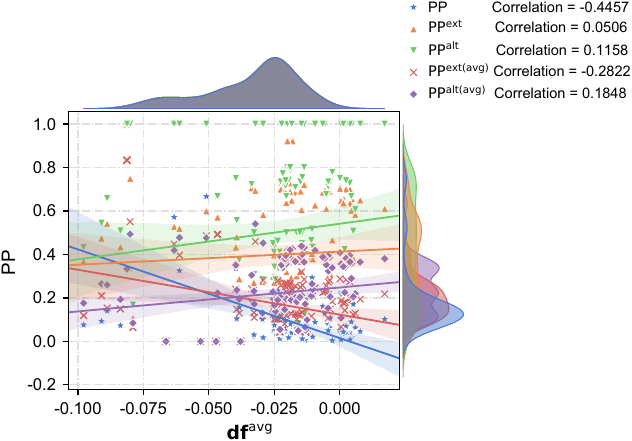}}
\vspace{-1mm}\caption{\small
Continuation of Figure~\ref{fig:reln}, that is, relation between individual fairness and group fairness. 
(a--c) and (d--f) Using GEI and the average HFM \citep{bian2024approximating} as the individual fairness, respectively. 
}\label{fig:rel,contd1}
\end{figure}

\end{appendices}

\bibliography{nus_title_iso4,refsmac}


\end{document}

%% file: prefix.tex
\usepackage{xfrac}
\usepackage{bm}
\usepackage{booktabs}
\let\cline\cmidrule
\usepackage{lscape}
\usepackage{subfig}
\usepackage{cleveref}

\def\thres{\text{threshold}}
\def\indicator{\mathbb{I}}

\def\reqdesign{indirectly}
\def\poss{poss}
\def\senatt{sensitive attribute}%
\def\sapl{sensitive attributes}%

\def\ie{i.e.}
\def\eg{e.g.}
\def\aka{aka.}

\def\wrt{w.r.t.}
\def\iff{iff}

\newcommand{\defineq}{\triangleq}

\def\xneg{\mathbf{x}}
\def\xpos{\mathbf{a}}
\def\xqtb{\tilde{\bm{a}}}
\def\probD{\mathsf{D}}
\def\probP{\mathbb{P}}
\def\probE{\mathbb{E}}

\def\cffU{\mathsf{U}}
\def\cffV{\mathsf{V}}
\def\cffF{\mathsf{F}}
\def\cffZ{\mathsf{Z}}
\def\cffPA{\mathrm{pa}_i}
\def\cffIV{\mathsf{V}_i}

\def\dist{\bm{d}}
\def\acc{\mathrm{acc}}
\def\err{\mathrm{err}}
\def\disf{\bm{g}}

\def\newy{\ddot{y}}

\def\dfprev{\mathbf{df}_\text{prev}}
\def\dfpreva{\hat{\mathbf{df}}_\text{prev}}

\newcommand\indep{\protect\mathpalette{\protect\independent}{\perp}}
\def\independent#1#2{\mathrel{\rlap{$#1#2$}\mkern2mu{#1#2}}}

\newcommand{\tabincell}[2]{\begin{tabular}{@{}#1@{}}#2\end{tabular}}

\newcommand{\topequation}{%
  \setlength\abovedisplayskip{2pt}%
  \setlength\belowdisplayskip{2pt}%
  \small%
}

%% file: main_arxiv.bbl

\begin{thebibliography}{63}
\ifx \bisbn   \undefined \def \bisbn  #1{ISBN #1}\fi
\ifx \binits  \undefined \def \binits#1{#1}\fi
\ifx \bauthor  \undefined \def \bauthor#1{#1}\fi
\ifx \batitle  \undefined \def \batitle#1{#1}\fi
\ifx \bjtitle  \undefined \def \bjtitle#1{#1}\fi
\ifx \bvolume  \undefined \def \bvolume#1{\textbf{#1}}\fi
\ifx \byear  \undefined \def \byear#1{#1}\fi
\ifx \bissue  \undefined \def \bissue#1{#1}\fi
\ifx \bfpage  \undefined \def \bfpage#1{#1}\fi
\ifx \blpage  \undefined \def \blpage #1{#1}\fi
\ifx \burl  \undefined \def \burl#1{\textsf{#1}}\fi
\ifx \doiurl  \undefined \def \doiurl#1{\url{https://doi.org/#1}}\fi
\ifx \betal  \undefined \def \betal{\textit{et al.}}\fi
\ifx \binstitute  \undefined \def \binstitute#1{#1}\fi
\ifx \binstitutionaled  \undefined \def \binstitutionaled#1{#1}\fi
\ifx \bctitle  \undefined \def \bctitle#1{#1}\fi
\ifx \beditor  \undefined \def \beditor#1{#1}\fi
\ifx \bpublisher  \undefined \def \bpublisher#1{#1}\fi
\ifx \bbtitle  \undefined \def \bbtitle#1{#1}\fi
\ifx \bedition  \undefined \def \bedition#1{#1}\fi
\ifx \bseriesno  \undefined \def \bseriesno#1{#1}\fi
\ifx \blocation  \undefined \def \blocation#1{#1}\fi
\ifx \bsertitle  \undefined \def \bsertitle#1{#1}\fi
\ifx \bsnm \undefined \def \bsnm#1{#1}\fi
\ifx \bsuffix \undefined \def \bsuffix#1{#1}\fi
\ifx \bparticle \undefined \def \bparticle#1{#1}\fi
\ifx \barticle \undefined \def \barticle#1{#1}\fi
\bibcommenthead
\ifx \bconfdate \undefined \def \bconfdate #1{#1}\fi
\ifx \botherref \undefined \def \botherref #1{#1}\fi
\ifx \url \undefined \def \url#1{\textsf{#1}}\fi
\ifx \bchapter \undefined \def \bchapter#1{#1}\fi
\ifx \bbook \undefined \def \bbook#1{#1}\fi
\ifx \bcomment \undefined \def \bcomment#1{#1}\fi
\ifx \oauthor \undefined \def \oauthor#1{#1}\fi
\ifx \citeauthoryear \undefined \def \citeauthoryear#1{#1}\fi
\ifx \endbibitem  \undefined \def \endbibitem {}\fi
\ifx \bconflocation  \undefined \def \bconflocation#1{#1}\fi
\ifx \arxivurl  \undefined \def \arxivurl#1{\textsf{#1}}\fi
\csname PreBibitemsHook\endcsname

\bibitem[\protect\citeauthoryear{Obermeyer
  et~al.}{2019}]{obermeyer2019dissecting}
\begin{barticle}
\bauthor{\bsnm{Obermeyer}, \binits{Z.}},
\bauthor{\bsnm{Powers}, \binits{B.}},
\bauthor{\bsnm{Vogeli}, \binits{C.}},
\bauthor{\bsnm{Mullainathan}, \binits{S.}}:
\batitle{Dissecting racial bias in an algorithm used to manage the health of
  populations}.
\bjtitle{Science}
\bvolume{366}(\bissue{6464}),
\bfpage{447}--\blpage{453}
(\byear{2019})
\doiurl{10.1126/science.aax2342}
\end{barticle}
\endbibitem

\bibitem[\protect\citeauthoryear{Vlasceanu and
  Amodio}{2022}]{vlasceanu2022propagation}
\begin{barticle}
\bauthor{\bsnm{Vlasceanu}, \binits{M.}},
\bauthor{\bsnm{Amodio}, \binits{D.M.}}:
\batitle{Propagation of societal gender inequality by internet search
  algorithms}.
\bjtitle{Proc Natl Acad Sci U.S.A.}
\bvolume{119}(\bissue{29}),
\bfpage{2204529119}
(\byear{2022})
\end{barticle}
\endbibitem

\bibitem[\protect\citeauthoryear{Chen et~al.}{2023}]{chen2023algorithmic}
\begin{barticle}
\bauthor{\bsnm{Chen}, \binits{R.J.}},
\bauthor{\bsnm{Wang}, \binits{J.J.}},
\bauthor{\bsnm{Williamson}, \binits{D.F.}},
\bauthor{\bsnm{Chen}, \binits{T.Y.}},
\bauthor{\bsnm{Lipkova}, \binits{J.}},
\bauthor{\bsnm{Lu}, \binits{M.Y.}},
\bauthor{\bsnm{Sahai}, \binits{S.}},
\bauthor{\bsnm{Mahmood}, \binits{F.}}:
\batitle{Algorithmic fairness in artificial intelligence for medicine and
  healthcare}.
\bjtitle{Nat Biomed Eng}
\bvolume{7}(\bissue{6}),
\bfpage{719}--\blpage{742}
(\byear{2023})
\end{barticle}
\endbibitem

\bibitem[\protect\citeauthoryear{Hu et~al.}{2024}]{hu2024generative}
\begin{botherref}
\oauthor{\bsnm{Hu}, \binits{T.}},
\oauthor{\bsnm{Kyrychenko}, \binits{Y.}},
\oauthor{\bsnm{Rathje}, \binits{S.}},
\oauthor{\bsnm{Collier}, \binits{N.}},
\oauthor{\bsnm{Linden}, \binits{S.}},
\oauthor{\bsnm{Roozenbeek}, \binits{J.}}:
Generative language models exhibit social identity biases.
Nat Comput Sci,
1--11
(2024)
\end{botherref}
\endbibitem

\bibitem[\protect\citeauthoryear{Glickman and Sharot}{2024}]{glickman2024human}
\begin{botherref}
\oauthor{\bsnm{Glickman}, \binits{M.}},
\oauthor{\bsnm{Sharot}, \binits{T.}}:
How human-ai feedback loops alter human perceptual, emotional and social
  judgements.
Nat Hum Behav
(2024)
\end{botherref}
\endbibitem

\bibitem[\protect\citeauthoryear{Jones et~al.}{2024}]{jones2024causal}
\begin{barticle}
\bauthor{\bsnm{Jones}, \binits{C.}},
\bauthor{\bsnm{Castro}, \binits{D.C.}},
\bauthor{\bsnm{De~Sousa~Ribeiro}, \binits{F.}},
\bauthor{\bsnm{Oktay}, \binits{O.}},
\bauthor{\bsnm{McCradden}, \binits{M.}},
\bauthor{\bsnm{Glocker}, \binits{B.}}:
\batitle{A causal perspective on dataset bias in machine learning for medical
  imaging}.
\bjtitle{Nat Mach Intell}
\bvolume{6}(\bissue{2}),
\bfpage{138}--\blpage{146}
(\byear{2024})
\end{barticle}
\endbibitem

\bibitem[\protect\citeauthoryear{Rampisela et~al.}{2025}]{rampisela2025joint}
\begin{bchapter}
\bauthor{\bsnm{Rampisela}, \binits{T.V.}},
\bauthor{\bsnm{Ruotsalo}, \binits{T.}},
\bauthor{\bsnm{Maistro}, \binits{M.}},
\bauthor{\bsnm{Lioma}, \binits{C.}}:
\bctitle{Joint evaluation of fairness and relevance in recommender systems with
  pareto frontier}.
In: \bbtitle{WWW},
pp. \bfpage{1548}--\blpage{1566}
(\byear{2025})
\end{bchapter}
\endbibitem

\bibitem[\protect\citeauthoryear{Binns}{2018}]{binns2018fairness}
\begin{bchapter}
\bauthor{\bsnm{Binns}, \binits{R.}}:
\bctitle{Fairness in machine learning: Lessons from political philosophy}.
In: \bbtitle{FAT},
pp. \bfpage{149}--\blpage{159}
(\byear{2018}).
\bcomment{PMLR}
\end{bchapter}
\endbibitem

\bibitem[\protect\citeauthoryear{Caton and Haas}{2020}]{caton2020fairness}
\begin{botherref}
\oauthor{\bsnm{Caton}, \binits{S.}},
\oauthor{\bsnm{Haas}, \binits{C.}}:
Fairness in machine learning: A survey.
ACM Comput Surv
(2020)
\end{botherref}
\endbibitem

\bibitem[\protect\citeauthoryear{Mitchell
  et~al.}{2021}]{mitchell2021algorithmic}
\begin{barticle}
\bauthor{\bsnm{Mitchell}, \binits{S.}},
\bauthor{\bsnm{Potash}, \binits{E.}},
\bauthor{\bsnm{Barocas}, \binits{S.}},
\bauthor{\bsnm{D'Amour}, \binits{A.}},
\bauthor{\bsnm{Lum}, \binits{K.}}:
\batitle{Algorithmic fairness: Choices, assumptions, and definitions}.
\bjtitle{Annu Rev Stat Appl}
\bvolume{8},
\bfpage{141}--\blpage{163}
(\byear{2021})
\end{barticle}
\endbibitem

\bibitem[\protect\citeauthoryear{Mehrabi et~al.}{2021}]{mehrabi2021survey}
\begin{barticle}
\bauthor{\bsnm{Mehrabi}, \binits{N.}},
\bauthor{\bsnm{Morstatter}, \binits{F.}},
\bauthor{\bsnm{Saxena}, \binits{N.}},
\bauthor{\bsnm{Lerman}, \binits{K.}},
\bauthor{\bsnm{Galstyan}, \binits{A.}}:
\batitle{A survey on bias and fairness in machine learning}.
\bjtitle{ACM Comput Surv}
\bvolume{54}(\bissue{6}),
\bfpage{1}--\blpage{35}
(\byear{2021})
\end{barticle}
\endbibitem

\bibitem[\protect\citeauthoryear{Pessach and Shmueli}{2022}]{pessach2022review}
\begin{barticle}
\bauthor{\bsnm{Pessach}, \binits{D.}},
\bauthor{\bsnm{Shmueli}, \binits{E.}}:
\batitle{A review on fairness in machine learning}.
\bjtitle{ACM Comput Surv}
\bvolume{55}(\bissue{3}),
\bfpage{1}--\blpage{44}
(\byear{2022})
\end{barticle}
\endbibitem

\bibitem[\protect\citeauthoryear{Ferrara}{2023}]{ferrara2023fairness}
\begin{barticle}
\bauthor{\bsnm{Ferrara}, \binits{E.}}:
\batitle{Fairness and bias in artificial intelligence: A brief survey of
  sources, impacts, and mitigation strategies}.
\bjtitle{Sci}
\bvolume{6}(\bissue{1}),
\bfpage{3}
(\byear{2023})
\end{barticle}
\endbibitem

\bibitem[\protect\citeauthoryear{Bertrand and Duflo}{2017}]{bertrand2017field}
\begin{bchapter}
\bauthor{\bsnm{Bertrand}, \binits{M.}},
\bauthor{\bsnm{Duflo}, \binits{E.}}:
\bctitle{Chapter 8 - field experiments on discrimination}.
In: \beditor{\bsnm{Banerjee}, \binits{A.V.}},
\beditor{\bsnm{Duflo}, \binits{E.}} (eds.)
\bbtitle{Handbook of Field Experiments}
vol. \bseriesno{1},
pp. \bfpage{309}--\blpage{393}.
\bpublisher{North-Holland},
\blocation{Amsterdam, Netherlands}
(\byear{2017}).
\doiurl{10.1016/bs.hefe.2016.08.004}
\end{bchapter}
\endbibitem

\bibitem[\protect\citeauthoryear{Barocas et~al.}{2023}]{barocas2023fairness}
\begin{bbook}
\bauthor{\bsnm{Barocas}, \binits{S.}},
\bauthor{\bsnm{Hardt}, \binits{M.}},
\bauthor{\bsnm{Narayanan}, \binits{A.}}:
\bbtitle{Fairness and Machine Learning: Limitations and Opportunities}.
\bpublisher{MIT Press},
\blocation{Cambridge, MA, USA}
(\byear{2023})
\end{bbook}
\endbibitem

\bibitem[\protect\citeauthoryear{Pager and Shepherd}{2008}]{pager2008sociology}
\begin{barticle}
\bauthor{\bsnm{Pager}, \binits{D.}},
\bauthor{\bsnm{Shepherd}, \binits{H.}}:
\batitle{The sociology of discrimination: Racial discrimination in employment,
  housing, credit, and consumer markets}.
\bjtitle{Annu Rev Sociol}
\bvolume{34}(\bissue{1}),
\bfpage{181}--\blpage{209}
(\byear{2008})
\end{barticle}
\endbibitem

\bibitem[\protect\citeauthoryear{Gajane and
  Pechenizkiy}{2018}]{gajane2017formalizing}
\begin{bchapter}
\bauthor{\bsnm{Gajane}, \binits{P.}},
\bauthor{\bsnm{Pechenizkiy}, \binits{M.}}:
\bctitle{On formalizing fairness in prediction with machine learning}.
In: \bbtitle{FAT/ML}
(\byear{2018})
\end{bchapter}
\endbibitem

\bibitem[\protect\citeauthoryear{Jiang et~al.}{2020}]{jiang2020wasserstein}
\begin{bchapter}
\bauthor{\bsnm{Jiang}, \binits{R.}},
\bauthor{\bsnm{Pacchiano}, \binits{A.}},
\bauthor{\bsnm{Stepleton}, \binits{T.}},
\bauthor{\bsnm{Jiang}, \binits{H.}},
\bauthor{\bsnm{Chiappa}, \binits{S.}}:
\bctitle{Wasserstein fair classification}.
In: \bbtitle{UAI},
pp. \bfpage{862}--\blpage{872}
(\byear{2020}).
\bcomment{PMLR}
\end{bchapter}
\endbibitem

\bibitem[\protect\citeauthoryear{Dwork et~al.}{2012}]{dwork2012fairness}
\begin{bchapter}
\bauthor{\bsnm{Dwork}, \binits{C.}},
\bauthor{\bsnm{Hardt}, \binits{M.}},
\bauthor{\bsnm{Pitassi}, \binits{T.}},
\bauthor{\bsnm{Reingold}, \binits{O.}},
\bauthor{\bsnm{Zemel}, \binits{R.}}:
\bctitle{Fairness through awareness}.
In: \bbtitle{ITCS}.
\bsertitle{ITCS '12},
pp. \bfpage{214}--\blpage{226}.
\bpublisher{ACM},
\blocation{New York, NY, USA}
(\byear{2012})
\end{bchapter}
\endbibitem

\bibitem[\protect\citeauthoryear{Chouldechova}{2017}]{chouldechova2017fair}
\begin{barticle}
\bauthor{\bsnm{Chouldechova}, \binits{A.}}:
\batitle{Fair prediction with disparate impact: A study of bias in recidivism
  prediction instruments}.
\bjtitle{Big Data}
\bvolume{5}(\bissue{2}),
\bfpage{153}--\blpage{163}
(\byear{2017})
\end{barticle}
\endbibitem

\bibitem[\protect\citeauthoryear{Zafar et~al.}{2017}]{zafar2017fairness2}
\begin{bchapter}
\bauthor{\bsnm{Zafar}, \binits{M.B.}},
\bauthor{\bsnm{Valera}, \binits{I.}},
\bauthor{\bsnm{Gomez~Rodriguez}, \binits{M.}},
\bauthor{\bsnm{Gummadi}, \binits{K.P.}}:
\bctitle{Fairness beyond disparate treatment \& disparate impact: Learning
  classification without disparate mistreatment}.
In: \bbtitle{WWW},
pp. \bfpage{1171}--\blpage{1180}
(\byear{2017})
\end{bchapter}
\endbibitem

\bibitem[\protect\citeauthoryear{Corbett-Davies
  et~al.}{2017}]{corbett2017algorithmic}
\begin{bchapter}
\bauthor{\bsnm{Corbett-Davies}, \binits{S.}},
\bauthor{\bsnm{Pierson}, \binits{E.}},
\bauthor{\bsnm{Feller}, \binits{A.}},
\bauthor{\bsnm{Goel}, \binits{S.}},
\bauthor{\bsnm{Huq}, \binits{A.}}:
\bctitle{Algorithmic decision making and the cost of fairness}.
In: \bbtitle{SIGKDD},
pp. \bfpage{797}--\blpage{806}.
\bpublisher{ACM},
\blocation{New York, NY, USA}
(\byear{2017})
\end{bchapter}
\endbibitem

\bibitem[\protect\citeauthoryear{Haas}{2019}]{haas2019price}
\begin{bchapter}
\bauthor{\bsnm{Haas}, \binits{C.}}:
\bctitle{The price of fairness - a framework to explore trade-offs in
  algorithmic fairness}.
In: \bbtitle{ICIS}
(\byear{2019}).
\bcomment{Association for Information Systems}
\end{bchapter}
\endbibitem

\bibitem[\protect\citeauthoryear{Agarwal et~al.}{2019}]{agarwal2019fair}
\begin{bchapter}
\bauthor{\bsnm{Agarwal}, \binits{A.}},
\bauthor{\bsnm{Dud{\'\i}k}, \binits{M.}},
\bauthor{\bsnm{Wu}, \binits{Z.S.}}:
\bctitle{Fair regression: Quantitative definitions and reduction-based
  algorithms}.
In: \bbtitle{ICML},
vol. \bseriesno{97},
pp. \bfpage{120}--\blpage{129}
(\byear{2019}).
\bcomment{PMLR}
\end{bchapter}
\endbibitem

\bibitem[\protect\citeauthoryear{Chen et~al.}{2024}]{chen2024fairness}
\begin{bchapter}
\bauthor{\bsnm{Chen}, \binits{Z.}},
\bauthor{\bsnm{Zhang}, \binits{J.M.}},
\bauthor{\bsnm{Sarro}, \binits{F.}},
\bauthor{\bsnm{Harman}, \binits{M.}}:
\bctitle{Fairness improvement with multiple protected attributes: How far are
  we?}
In: \bbtitle{ICSE},
pp. \bfpage{1}--\blpage{13}
(\byear{2024})
\end{bchapter}
\endbibitem

\bibitem[\protect\citeauthoryear{Feldman et~al.}{2015}]{feldman2015certifying}
\begin{bchapter}
\bauthor{\bsnm{Feldman}, \binits{M.}},
\bauthor{\bsnm{Friedler}, \binits{S.A.}},
\bauthor{\bsnm{Moeller}, \binits{J.}},
\bauthor{\bsnm{Scheidegger}, \binits{C.}},
\bauthor{\bsnm{Venkatasubramanian}, \binits{S.}}:
\bctitle{Certifying and removing disparate impact}.
In: \bbtitle{SIGKDD},
pp. \bfpage{259}--\blpage{268}
(\byear{2015})
\end{bchapter}
\endbibitem

\bibitem[\protect\citeauthoryear{Diana et~al.}{2024}]{diana2024minimax}
\begin{bchapter}
\bauthor{\bsnm{Diana}, \binits{E.}},
\bauthor{\bsnm{Sharifi-Malvajerdi}, \binits{S.}},
\bauthor{\bsnm{Vakilian}, \binits{A.}}:
\bctitle{Minimax group fairness in strategic classification}.
In: \bbtitle{SatML}
(\byear{2024})
\end{bchapter}
\endbibitem

\bibitem[\protect\citeauthoryear{Hardt et~al.}{2016}]{hardt2016equality}
\begin{bchapter}
\bauthor{\bsnm{Hardt}, \binits{M.}},
\bauthor{\bsnm{Price}, \binits{E.}},
\bauthor{\bsnm{Srebro}, \binits{N.}}:
\bctitle{Equality of opportunity in supervised learning}.
In: \beditor{\bsnm{Lee}, \binits{D.}},
\beditor{\bsnm{Sugiyama}, \binits{M.}},
\beditor{\bsnm{Luxburg}, \binits{U.}},
\beditor{\bsnm{Guyon}, \binits{I.}},
\beditor{\bsnm{Garnett}, \binits{R.}} (eds.)
\bbtitle{NIPS},
vol. \bseriesno{29},
pp. \bfpage{3323}--\blpage{3331}.
\bpublisher{Curran Associates Inc.},
\blocation{Red Hook, NY, USA}
(\byear{2016})
\end{bchapter}
\endbibitem

\bibitem[\protect\citeauthoryear{Kearns et~al.}{2018}]{kearns2018preventing}
\begin{bchapter}
\bauthor{\bsnm{Kearns}, \binits{M.}},
\bauthor{\bsnm{Neel}, \binits{S.}},
\bauthor{\bsnm{Roth}, \binits{A.}},
\bauthor{\bsnm{Wu}, \binits{Z.S.}}:
\bctitle{Preventing fairness gerrymandering: Auditing and learning for subgroup
  fairness}.
In: \bbtitle{ICML},
pp. \bfpage{2564}--\blpage{2572}
(\byear{2018}).
\bcomment{PMLR}
\end{bchapter}
\endbibitem

\bibitem[\protect\citeauthoryear{Kearns et~al.}{2019}]{kearns2019empirical}
\begin{bchapter}
\bauthor{\bsnm{Kearns}, \binits{M.}},
\bauthor{\bsnm{Neel}, \binits{S.}},
\bauthor{\bsnm{Roth}, \binits{A.}},
\bauthor{\bsnm{Wu}, \binits{Z.S.}}:
\bctitle{An empirical study of rich subgroup fairness for machine learning}.
In: \bbtitle{FAT},
pp. \bfpage{100}--\blpage{109}
(\byear{2019})
\end{bchapter}
\endbibitem

\bibitem[\protect\citeauthoryear{Verma and Rubin}{2018}]{verma2018fairness}
\begin{bchapter}
\bauthor{\bsnm{Verma}, \binits{S.}},
\bauthor{\bsnm{Rubin}, \binits{J.}}:
\bctitle{Fairness definitions explained}.
In: \bbtitle{FairWare},
pp. \bfpage{1}--\blpage{7}
(\byear{2018})
\end{bchapter}
\endbibitem

\bibitem[\protect\citeauthoryear{Speicher et~al.}{2018}]{speicher2018unified}
\begin{bchapter}
\bauthor{\bsnm{Speicher}, \binits{T.}},
\bauthor{\bsnm{Heidari}, \binits{H.}},
\bauthor{\bsnm{Grgic-Hlaca}, \binits{N.}},
\bauthor{\bsnm{Gummadi}, \binits{K.P.}},
\bauthor{\bsnm{Singla}, \binits{A.}},
\bauthor{\bsnm{Weller}, \binits{A.}},
\bauthor{\bsnm{Zafar}, \binits{M.B.}}:
\bctitle{A unified approach to quantifying algorithmic unfairness: Measuring
  individual \&group unfairness via inequality indices}.
In: \bbtitle{SIGKDD},
pp. \bfpage{2239}--\blpage{2248}
(\byear{2018})
\end{bchapter}
\endbibitem

\bibitem[\protect\citeauthoryear{Kusner
  et~al.}{2017}]{kusner2017counterfactual}
\begin{bchapter}
\bauthor{\bsnm{Kusner}, \binits{M.J.}},
\bauthor{\bsnm{Loftus}, \binits{J.}},
\bauthor{\bsnm{Russell}, \binits{C.}},
\bauthor{\bsnm{Silva}, \binits{R.}}:
\bctitle{Counterfactual fairness}.
In: \bbtitle{NIPS},
vol. \bseriesno{30},
pp. \bfpage{4069}--\blpage{4079}
(\byear{2017}).
\bcomment{NIPS Proceedings}
\end{bchapter}
\endbibitem

\bibitem[\protect\citeauthoryear{Kilbertus
  et~al.}{2017}]{kilbertus2017avoiding}
\begin{bchapter}
\bauthor{\bsnm{Kilbertus}, \binits{N.}},
\bauthor{\bsnm{Rojas~Carulla}, \binits{M.}},
\bauthor{\bsnm{Parascandolo}, \binits{G.}},
\bauthor{\bsnm{Hardt}, \binits{M.}},
\bauthor{\bsnm{Janzing}, \binits{D.}},
\bauthor{\bsnm{Sch{\"o}lkopf}, \binits{B.}}:
\bctitle{Avoiding discrimination through causal reasoning}.
In: \bbtitle{NIPS},
vol. \bseriesno{30}
(\byear{2017})
\end{bchapter}
\endbibitem

\bibitem[\protect\citeauthoryear{Bian and Zhang}{2023}]{bian2023increasing_alt}
\begin{botherref}
\oauthor{\bsnm{Bian}, \binits{Y.}},
\oauthor{\bsnm{Zhang}, \binits{K.}}:
Increasing fairness via combination with learning guarantees.
arXiv preprint arXiv:2301.10813
(2023)
\end{botherref}
\endbibitem

\bibitem[\protect\citeauthoryear{Kim et~al.}{2019}]{kim2019multiaccuracy}
\begin{bchapter}
\bauthor{\bsnm{Kim}, \binits{M.P.}},
\bauthor{\bsnm{Ghorbani}, \binits{A.}},
\bauthor{\bsnm{Zou}, \binits{J.}}:
\bctitle{Multiaccuracy: Black-box post-processing for fairness in
  classification}.
In: \bbtitle{AIES}.
\bsertitle{AIES '19},
pp. \bfpage{247}--\blpage{254}.
\bpublisher{Association for Computing Machinery},
\blocation{New York, NY, USA}
(\byear{2019})
\end{bchapter}
\endbibitem

\bibitem[\protect\citeauthoryear{Foulds
  et~al.}{2020}]{foulds2020intersectional}
\begin{bchapter}
\bauthor{\bsnm{Foulds}, \binits{J.R.}},
\bauthor{\bsnm{Islam}, \binits{R.}},
\bauthor{\bsnm{Keya}, \binits{K.N.}},
\bauthor{\bsnm{Pan}, \binits{S.}}:
\bctitle{An intersectional definition of fairness}.
In: \bbtitle{ICDE},
pp. \bfpage{1918}--\blpage{1921}
(\byear{2020}).
\bcomment{IEEE}
\end{bchapter}
\endbibitem

\bibitem[\protect\citeauthoryear{Ghosh et~al.}{2021}]{ghosh2021characterizing}
\begin{bchapter}
\bauthor{\bsnm{Ghosh}, \binits{A.}},
\bauthor{\bsnm{Genuit}, \binits{L.}},
\bauthor{\bsnm{Reagan}, \binits{M.}}:
\bctitle{Characterizing intersectional group fairness with worst-case
  comparisons}.
In: \beditor{\bsnm{Lamba}, \binits{D.}},
\beditor{\bsnm{Hsu}, \binits{W.H.}} (eds.)
\bbtitle{Workshop on AIDBEI},
vol. \bseriesno{142}.
\bconflocation{Virtual},
pp. \bfpage{22}--\blpage{34}
(\byear{2021}).
\bcomment{PMLR}
\end{bchapter}
\endbibitem

\bibitem[\protect\citeauthoryear{Grgi{\'c}-Hla{\v{c}}a
  et~al.}{2016}]{grgic2016case}
\begin{bchapter}
\bauthor{\bsnm{Grgi{\'c}-Hla{\v{c}}a}, \binits{N.}},
\bauthor{\bsnm{Zafar}, \binits{M.B.}},
\bauthor{\bsnm{Gummadi}, \binits{K.P.}},
\bauthor{\bsnm{Weller}, \binits{A.}}:
\bctitle{The case for process fairness in learning: Feature selection for fair
  decision making}.
In: \bbtitle{NIPS Symposium on Machine Learning and the Law},
vol. \bseriesno{1},
p. \bfpage{11}
(\byear{2016}).
\bcomment{Barcelona, Spain}
\end{bchapter}
\endbibitem

\bibitem[\protect\citeauthoryear{Grgi{\'c}-Hla{\v{c}}a
  et~al.}{2018}]{grgic2018beyond}
\begin{bchapter}
\bauthor{\bsnm{Grgi{\'c}-Hla{\v{c}}a}, \binits{N.}},
\bauthor{\bsnm{Zafar}, \binits{M.B.}},
\bauthor{\bsnm{Gummadi}, \binits{K.P.}},
\bauthor{\bsnm{Weller}, \binits{A.}}:
\bctitle{Beyond distributive fairness in algorithmic decision making: Feature
  selection for procedurally fair learning}.
In: \bbtitle{AAAI},
vol. \bseriesno{32}
(\byear{2018})
\end{bchapter}
\endbibitem

\bibitem[\protect\citeauthoryear{Wang et~al.}{2024}]{wang2024procedural}
\begin{botherref}
\oauthor{\bsnm{Wang}, \binits{Z.}},
\oauthor{\bsnm{Huang}, \binits{C.}},
\oauthor{\bsnm{Yao}, \binits{X.}}:
Procedural fairness in machine learning.
arXiv preprint arXiv:2404.01877
(2024)
\end{botherref}
\endbibitem

\bibitem[\protect\citeauthoryear{Bian and Luo}{2024}]{bian2024does}
\begin{botherref}
\oauthor{\bsnm{Bian}, \binits{Y.}},
\oauthor{\bsnm{Luo}, \binits{Y.}}:
Does machine bring in extra bias in learning? approximating fairness in models
  promptly.
arXiv preprint arXiv:2405.09251
(2024)
\end{botherref}
\endbibitem

\bibitem[\protect\citeauthoryear{Bian et~al.}{2024}]{bian2024approximating}
\begin{botherref}
\oauthor{\bsnm{Bian}, \binits{Y.}},
\oauthor{\bsnm{Luo}, \binits{Y.}},
\oauthor{\bsnm{Xu}, \binits{P.}}:
Approximating discrimination within models when faced with several non-binary
  sensitive attributes.
arXiv preprint arXiv:2408.06099
(2024).
Under review
\end{botherref}
\endbibitem

\bibitem[\protect\citeauthoryear{Akosa}{2017}]{akosa2017predictive}
\begin{bchapter}
\bauthor{\bsnm{Akosa}, \binits{J.}}:
\bctitle{Predictive accuracy: A misleading performance measure for highly
  imbalanced data}.
In: \bbtitle{Proceedings of the SAS Global Forum},
vol. \bseriesno{12},
pp. \bfpage{1}--\blpage{4}
(\byear{2017}).
\bcomment{SAS Institute Inc. Cary, NC, USA}
\end{bchapter}
\endbibitem

\bibitem[\protect\citeauthoryear{Berk et~al.}{2021}]{berk2021fairness}
\begin{barticle}
\bauthor{\bsnm{Berk}, \binits{R.}},
\bauthor{\bsnm{Heidari}, \binits{H.}},
\bauthor{\bsnm{Jabbari}, \binits{S.}},
\bauthor{\bsnm{Kearns}, \binits{M.}},
\bauthor{\bsnm{Roth}, \binits{A.}}:
\batitle{Fairness in criminal justice risk assessments: The state of the art}.
\bjtitle{Sociol Methods Res}
\bvolume{50}(\bissue{1}),
\bfpage{3}--\blpage{44}
(\byear{2021})
\end{barticle}
\endbibitem

\bibitem[\protect\citeauthoryear{Pleiss et~al.}{2017}]{pleiss2017fairness}
\begin{bchapter}
\bauthor{\bsnm{Pleiss}, \binits{G.}},
\bauthor{\bsnm{Raghavan}, \binits{M.}},
\bauthor{\bsnm{Wu}, \binits{F.}},
\bauthor{\bsnm{Kleinberg}, \binits{J.}},
\bauthor{\bsnm{Weinberger}, \binits{K.Q.}}:
\bctitle{On fairness and calibration}.
In: \bbtitle{NIPS},
vol. \bseriesno{30}
(\byear{2017})
\end{bchapter}
\endbibitem

\bibitem[\protect\citeauthoryear{Zliobaite}{2015}]{zliobaite2015relation}
\begin{bchapter}
\bauthor{\bsnm{Zliobaite}, \binits{I.}}:
\bctitle{On the relation between accuracy and fairness in binary
  classification}.
In: \bbtitle{ICML Workshop on FATML}
(\byear{2015})
\end{bchapter}
\endbibitem

\bibitem[\protect\citeauthoryear{Simoiu et~al.}{2017}]{simoiu2017problem}
\begin{barticle}
\bauthor{\bsnm{Simoiu}, \binits{C.}},
\bauthor{\bsnm{Corbett-Davies}, \binits{S.}},
\bauthor{\bsnm{Goel}, \binits{S.}}:
\batitle{The problem of infra-marginality in outcome tests for discrimination}.
\bjtitle{Ann Appl Stat}
\bvolume{11}(\bissue{3}),
\bfpage{1193}--\blpage{1216}
(\byear{2017})
\end{barticle}
\endbibitem

\bibitem[\protect\citeauthoryear{Goel et~al.}{2016}]{goel2016precinct}
\begin{barticle}
\bauthor{\bsnm{Goel}, \binits{S.}},
\bauthor{\bsnm{Rao}, \binits{J.M.}},
\bauthor{\bsnm{Shroff}, \binits{R.}}:
\batitle{Precinct or prejudice? understanding racial disparities in new york
  city's stop-and-frisk policy}.
\bjtitle{Ann Appl Stat}
\bvolume{10}(\bissue{1}),
\bfpage{365}--\blpage{394}
(\byear{2016})
\end{barticle}
\endbibitem

\bibitem[\protect\citeauthoryear{Luong et~al.}{2011}]{luong2011k}
\begin{bchapter}
\bauthor{\bsnm{Luong}, \binits{B.T.}},
\bauthor{\bsnm{Ruggieri}, \binits{S.}},
\bauthor{\bsnm{Turini}, \binits{F.}}:
\bctitle{k-nn as an implementation of situation testing for discrimination
  discovery and prevention}.
In: \bbtitle{SIGKDD},
pp. \bfpage{502}--\blpage{510}
(\byear{2011})
\end{bchapter}
\endbibitem

\bibitem[\protect\citeauthoryear{Boeschoten
  et~al.}{2021}]{boeschoten2021achieving}
\begin{botherref}
\oauthor{\bsnm{Boeschoten}, \binits{L.}},
\oauthor{\bsnm{Kesteren}, \binits{E.-J.}},
\oauthor{\bsnm{Bagheri}, \binits{A.}},
\oauthor{\bsnm{Oberski}, \binits{D.L.}}:
Achieving fair inference using error-prone outcomes.
Int J Interact Multimed Artif Intell
\textbf{6}(5)
(2021)
\end{botherref}
\endbibitem

\bibitem[\protect\citeauthoryear{Nabi and Shpitser}{2018}]{nabi2018fair}
\begin{bchapter}
\bauthor{\bsnm{Nabi}, \binits{R.}},
\bauthor{\bsnm{Shpitser}, \binits{I.}}:
\bctitle{Fair inference on outcomes}.
In: \bbtitle{AAAI},
vol. \bseriesno{32}
(\byear{2018})
\end{bchapter}
\endbibitem

\bibitem[\protect\citeauthoryear{Kleinberg
  et~al.}{2017}]{kleinberg2016inherent}
\begin{bchapter}
\bauthor{\bsnm{Kleinberg}, \binits{J.}},
\bauthor{\bsnm{Mullainathan}, \binits{S.}},
\bauthor{\bsnm{Raghavan}, \binits{M.}}:
\bctitle{Inherent trade-offs in the fair determination of risk scores}.
In: \bbtitle{ITCS}
(\byear{2017})
\end{bchapter}
\endbibitem

\bibitem[\protect\citeauthoryear{H{\'e}bert-Johnson
  et~al.}{2018}]{hebert2018multicalibration}
\begin{bchapter}
\bauthor{\bsnm{H{\'e}bert-Johnson}, \binits{U.}},
\bauthor{\bsnm{Kim}, \binits{M.}},
\bauthor{\bsnm{Reingold}, \binits{O.}},
\bauthor{\bsnm{Rothblum}, \binits{G.}}:
\bctitle{Multicalibration: Calibration for the (computationally-identifiable)
  masses}.
In: \bbtitle{ICML},
pp. \bfpage{1939}--\blpage{1948}
(\byear{2018}).
\bcomment{PMLR}
\end{bchapter}
\endbibitem

\bibitem[\protect\citeauthoryear{Gohar and Cheng}{2023}]{gohar2023survey}
\begin{bchapter}
\bauthor{\bsnm{Gohar}, \binits{U.}},
\bauthor{\bsnm{Cheng}, \binits{L.}}:
\bctitle{A survey on intersectional fairness in machine learning: notions,
  mitigation, and challenges}.
In: \bbtitle{IJCAI},
pp. \bfpage{6619}--\blpage{6627}
(\byear{2023})
\end{bchapter}
\endbibitem

\bibitem[\protect\citeauthoryear{Gartner}{2023}]{gartner2025new}
\begin{botherref}
\oauthor{\bsnm{Gartner}, \binits{J.}}:
A new metric for quantifying machine learning fairness in healthcare.
[EB/OL].
Latest accessed August 14, 2025
(2023).
\url{https://www.closedloop.ai/blog/a-new-metric-for-quantifying-machine-learning-fairness-in-healthcare/}
\end{botherref}
\endbibitem

\bibitem[\protect\citeauthoryear{Ricci}{}]{dataset1_2024}
\begin{botherref}
\oauthor{\bsnm{Ricci}}:
Ricci: Firefighter promotion exam scores.
\url{https://rdrr.io/cran/Stat2Data/man/Ricci.html}
\end{botherref}
\endbibitem

\bibitem[\protect\citeauthoryear{Credit}{}]{dataset2_2024}
\begin{botherref}
\oauthor{\bsnm{Credit}}:
Statlog (German credit data).
\url{https://archive.ics.uci.edu/dataset/144/statlog+german+credit+data}
\end{botherref}
\endbibitem

\bibitem[\protect\citeauthoryear{Income}{}]{dataset3_2024}
\begin{botherref}
\oauthor{\bsnm{Income}}:
Adult.
\url{https://archive.ics.uci.edu/dataset/2/adult}
\end{botherref}
\endbibitem

\bibitem[\protect\citeauthoryear{COMPAS}{}]{dataset4_2024}
\begin{botherref}
\oauthor{\bsnm{COMPAS}}:
Propublica-recidivism and Propublica-violent-recidivism datasets.
[EB/OL].
Latest accessed August 29, 2022.
\url{https://github.com/propublica/compas-analysis/}
\end{botherref}
\endbibitem

\bibitem[\protect\citeauthoryear{Ke et~al.}{2017}]{ke2017lightgbm}
\begin{bchapter}
\bauthor{\bsnm{Ke}, \binits{G.}},
\bauthor{\bsnm{Meng}, \binits{Q.}},
\bauthor{\bsnm{Finley}, \binits{T.}},
\bauthor{\bsnm{Wang}, \binits{T.}},
\bauthor{\bsnm{Chen}, \binits{W.}},
\bauthor{\bsnm{Ma}, \binits{W.}},
\bauthor{\bsnm{Ye}, \binits{Q.}},
\bauthor{\bsnm{Liu}, \binits{T.-Y.}}:
\bctitle{Lightgbm: A highly efficient gradient boosting decision tree}.
In: \bbtitle{NIPS},
vol. \bseriesno{30},
pp. \bfpage{3146}--\blpage{3154}
(\byear{2017})
\end{bchapter}
\endbibitem

\bibitem[\protect\citeauthoryear{Cruz et~al.}{2023}]{cruz2022fairgbm}
\begin{bchapter}
\bauthor{\bsnm{Cruz}, \binits{A.F.}},
\bauthor{\bsnm{Bel{\'e}m}, \binits{C.}},
\bauthor{\bsnm{Bravo}, \binits{J.}},
\bauthor{\bsnm{Saleiro}, \binits{P.}},
\bauthor{\bsnm{Bizarro}, \binits{P.}}:
\bctitle{Fairgbm: Gradient boosting with fairness constraints}.
In: \bbtitle{ICLR}
(\byear{2023})
\end{bchapter}
\endbibitem

\bibitem[\protect\citeauthoryear{Iosifidis and
  Ntoutsi}{2019}]{iosifidis2019adafair}
\begin{bchapter}
\bauthor{\bsnm{Iosifidis}, \binits{V.}},
\bauthor{\bsnm{Ntoutsi}, \binits{E.}}:
\bctitle{Adafair: Cumulative fairness adaptive boosting}.
In: \bbtitle{CIKM},
pp. \bfpage{781}--\blpage{790}.
\bpublisher{ACM},
\blocation{New York, NY, USA}
(\byear{2019})
\end{bchapter}
\endbibitem

\end{thebibliography}
